\documentclass[10pt]{article} 
\usepackage[accepted]{tmlr}

\usepackage{amsmath,amsfonts,bm}

\def\eqref#1{equation~\ref{#1}}

\def\1{\bm{1}}

\DeclareMathAlphabet{\mathsfit}{\encodingdefault}{\sfdefault}{m}{sl}
\SetMathAlphabet{\mathsfit}{bold}{\encodingdefault}{\sfdefault}{bx}{n}

\DeclareMathOperator*{\argmin}{arg\,min}

\usepackage{url}
\usepackage{graphicx}
\usepackage{multirow}
\usepackage{rotating}
\usepackage{amsmath}
\usepackage{amssymb}
\usepackage{array}
\usepackage{float}
\usepackage{color}
\usepackage{bbm}
\usepackage{breakurl}
\usepackage[unicode=true,pdfusetitle,
 bookmarks=true,bookmarksnumbered=false,bookmarksopen=false,
 breaklinks=true,pdfborder={0 0 0},pdfborderstyle={},backref=false,colorlinks=true]
 {hyperref}
\hypersetup{
 linkcolor=blue,citecolor=darkred,urlcolor=darkgreen}
\usepackage[ruled]{algorithm2e}

\definecolor{darkred}{rgb}{0.7, 0.0, 0.0}
\definecolor{darkgreen}{rgb}{0.05, 0.45, 0.27}

\title{Systematically and efficiently improving $k$-means initialization
by pairwise-nearest-neighbor smoothing}

\author{\name Carlo Baldassi \email carlo.baldassi@unibocconi.it \\
      \addr Department of Computing Sciences, ArtLab, BIDSA\\
      Bocconi University, Milan\\
      ELLIS Scholar
      }

\newcommand{\noun}[1]{\textsc{#1}}
\newcommand{\bfnoun}[1]{\textbf{\footnotesize{}\uppercase{#1}}}
\newcommand{\fdot}{.}

\def\month{12}  
\def\year{2022} 
\def\openreview{\url{https://openreview.net/forum?id=FTtFAg3pek}} 

\begin{document}

\maketitle

\begin{abstract}
We present a meta-method for initializing (seeding) the $k$-means
clustering algorithm called PNN-smoothing. It consists in splitting
a given dataset into $J$ random subsets, clustering each of them
individually, and merging the resulting clusterings with the pairwise-nearest-neighbor
(PNN) method. It is a meta-method in the sense that when clustering
the individual subsets any seeding algorithm can be used. If the computational
complexity of that seeding algorithm is linear in the size of the
data $N$ and the number of clusters $k$, PNN-smoothing is also almost
linear with an appropriate choice of $J$, and quite competitive in
practice. We show empirically, using several existing seeding methods
and testing on several synthetic and real datasets, that this procedure
results in systematically better costs. In particular, our method
of enhancing $k$-means++ seeding proves superior in both effectiveness
and speed compared to the popular ``greedy'' $k$-means++ variant. Our implementation
is publicly available at \href{https://github.com/carlobaldassi/KMeansPNNSmoothing.jl}{https://github.com/carlobaldassi/KMeansPNNSmoothing.jl}.
\end{abstract}

\section{Introduction\label{sec:intro}}

The classical $k$-means algorithm is one of the most well-known and
widely adopted clustering algorithms \citep{berkhin2006survey,wu2008top}.
Given $N$ data points $\mathcal{X}=\left(x_{i}\right)_{i=1..N}$,
where each point is $D$-dimensional, $x_{i}\in\mathbb{R}^{D}$, and
given an integer $k\ge2$, the algorithm aims at minimizing the sum-of-squared-errors
(SSE) cost, defined as a function of $k$ centroids $\mathcal{C}=\left(c_{a}\right)_{a=1..k}\in\left(\mathbb{R}^{D}\right)^{k}$,
and of a partition of the data points $\mathcal{P}=\left(p_{i}\right)_{i=1..N}\in\left\{ 1,\dots,k\right\} ^{N}$,
as such:

\begin{equation}
\mathrm{SSE}\left(\mathcal{C},\mathcal{P};\mathcal{X}\right)=\sum_{i=1}^{N}\left\Vert x_{i}-c_{p_{i}}\right\Vert ^{2}\label{eq:SSE}
\end{equation}

For fixed $\mathcal{C}$, the optimal partition is obtained by associating
each point to its nearest centroid, and conversely, for fixed $\mathcal{P}$,
the optimal centroids are given by the barycenter of each cluster.
The $k$-means algorithm starts from an initial guess for the configuration
and alternates optimizing $\mathcal{C}$ and $\mathcal{P}$ until
a fixed point is reached. This alternating procedure, due to Lloyd
\citep{lloyd1982least}, is greedy and converges to a local minimum.
Its computational cost is $O\left(kND\right)$ per iteration; both
optimization steps can be straightforwardly parallelized over $N$.
It is notoriously sensitive to the choice of the initial configuration,
i.e. the seeding, both in terms of the final value of the SSE and
of the number of Lloyd's iterations required to converge. Several
schemes have been proposed, with various degrees of complexity. An
extremely basic and cheap option is to sample the initial centroids
uniformly at random from $\mathcal{X}$ \citep{macqueen1967some}.
Other popular methods typically produce considerable improvement and
also scale like $O\left(kND\right)$, e.g. $k$-means++ \citep{arthur2007k}
and maxmin \citep{gonzalez1985clustering,katsavounidis1994new}, among
several others. Yet other methods have worse scalings and thus tend
to dominate the computational time, e.g. the pariwise-nearest-neighbors
(PNN) method \citep{equitz1989new} which scales like $\Omega\left(N^{2}\right)$.

In this paper, we propose a novel scheme, PNN-smoothing, based on
randomly splitting the dataset $\mathcal{X}$ into $J$ subsets, clustering
them individually with $k$-means, and then merging the resulting
$Jk$ clusters following the PNN procedure until only $k$ clusters
remain: these constitute the new seed for Lloyd's algorithm. It is
a meta-method, in the sense that it can use any seeding procedure
for the subsets. We denote this with \noun{pnns(init)} where \noun{init}
is any seeding algorithm. If \noun{init} is $O\left(kN\right)$, then
\noun{pnns(init)} is also almost linear, as long as we set $J=O\left(\sqrt{N/k}\right)$.
Our empirical tests indicate that \noun{pnns(init)} gives systematically
better SSEs compared to \noun{init}, at a very small computational
cost, even in a parallel implementation. Its results are surprisingly
good even when \noun{init} is one of the worst seeding methods, random
uniform initialization.

Throughout the paper, we focus exclusively on the effect of seeding
on basic $k$-means, which can be regarded as a basic tool in the
optimization of the $\mathrm{SSE}$ cost. Evaluating and contrasting
the effectiveness of different methods is not straightforward. Our
main metric will be the wall-clock computational time, measured using
a state-of-the-art implementation of all methods under uniform conditions.
It is also desirable, however, to provide some hardware- and language-independent
metric, beyond the asymptotic scaling analysis. It is often the case
that computing distances between vectors takes up the majority of
the computational effort and thus that the number of distance computations
correlates well with the running time, especially in high-dimensional
cases (although of course various circumstances, especially caching,
can significantly affect the computational cost of a distance computation
in practice) \citep{newling2016fast}. Indeed, several methods employed
to accelerate the Lloyd's iteration procedure are explicitly aimed
at reducing the number of distance computations (in the $\mathcal{P}$-from-$\mathcal{C}$
step) at the cost of some additional bookkeeping \citep{kaukoranta1999reduced,elkan2003using,hamerly2010making,ding2015yinyang,newling2016fast,xia2020fast}.

We thus define the number of normalized distance computations (NDC)
as the total number of distance (or squared-distance) computations
between $D$-dimensional vectors, wherever they may appear in a procedure,
divided by $Nk$. The normalization makes this measure comparable
across datasets, so that a value of $1$ corresponds to one computation
of the partition $\mathcal{P}$ from $\mathcal{C}$, performed from
scratch. Due to the variability in the optimization procedure, and
the use of accelerators, it is generally quite difficult to estimate
the NDC required by a given method. However, for some seeding algorithms
the number of NDC can be computed exactly, and for others it is rather
straightforward to at least provide some useful lower bounds. Note
that our implementation of Lloyd's algorithm starts by optimizing
$\mathcal{C}$ from $\mathcal{P}$, and thus we regard seeding algorithms
as producing an initial partition $\mathcal{P}$. This choice is justified
by the fact that several seeding algorithms, even if they are based
on picking some initial centroids, include the computation of $\mathcal{P}$
as a byproduct, and those that don't would need to perform this step
at least once in any case\footnote{We only consider the exact Lloyd's iterations here, in which case
sub-linear (in $N$) seeding methods such as \noun{af-kmc}$^{2}$
\citep{bachem2016fast} don't provide an advantage.}; therefore, this definition allows to compare all seeding algorithms
consistently. As a consequence, all the seeding algorithms that we
will consider in this work require at least $1$ NDC.

The rest of the paper is organized as follows. In sec.~\ref{sec:prior}
we review prior literature and describe a few seeding methods that
will be considered in the tests. In sec.~\ref{sec:pnns} we describe
and discuss in detail the \noun{pnns(init)} scheme. In sec.~\ref{sec:experiments}
we present and analyze detailed numerical results on several challenging
synthetic and non-synthetic datasets. Sec.~\ref{sec:discussion}
has a final discussion.

\section{Relation to prior works\label{sec:prior}}

As mentioned in the introduction, a large number of seeding schemes
for $k$-means have been proposed. Extensive reviews and benchmarks
can be found in refs.~\citep{celebi2013comparative,franti2019much}.
Here, we only cover a few selected ones, chosen on criteria of simplicity,
popularity, similarity with our scheme, and effectiveness. Their scaling
and NDCs characteristics (together with those of \noun{pnns(init)}
described in the next section) are also summarized in table~\ref{tab:summary-of-seeding}.

\bfnoun{unif.} Uniformly sampling (preferably without replacement)
$k$ centroids from the dataset \citep{macqueen1967some} is arguably
the most popular method. We'll call this seeding method \noun{unif}.
It's extremely simple, but its performance is generally very poor,
even in moderately hard circumstances, as it often leads to poor local
minima and long convergence times.

For the reasons explained in the
introduction, although the cost of sampling the centroids is $O\left(k\right)$,
the overall cost of \noun{unif} is still $O\left(kND\right)$, entirely
due to performing $1$ NDC.

\bfnoun{maxmin.} Another simple method \citep{gonzalez1985clustering,katsavounidis1994new}
goes under the name of ``furthest point'', or ``maxmin'', or ``maximin''.
We will refer to the same variant that was used in~\citet{celebi2013comparative,franti2019much},
and call it \noun{maxmin}. It consists in selecting the first centroid
at random from the dataset, after which the process is iterative and
deterministic: at each step, each successive centroid is chosen as
the furthest point from the centroids selected so far. More precisely,
it's the point that maximizes the distance from its nearest centroid:
$c_{a}=\mathrm{argmax}_{x\in\mathcal{X}}\min_{b<a}\left\Vert x-c_{b}\right\Vert ^{2}$.

This algorithm scales as $O\left(kND\right)$; it also computes the
optimal partition (with respect to the chosen centroids) as a byproduct
of the selection procedure: it requires precisely $1$ NDC. The results
of \citet{franti2019much}, on synthetic datasets, report this
method as being among the best of those that were tested. On the other
hand, in \citet{celebi2013comparative}, in which an array of
real datasets was also tested, the authors advise against this method.
They suggest instead, among the algorithms in the same complexity
class, to use greedy-$k$-means++ or Bradley and Fayyad's ``refine''
(both described below).

\bfnoun{{[}g{]}km++.} The ``$k$-means++'' seeding
method \citep{arthur2007k} can be regarded as a more stochastic version
of \noun{maxmin}. While the first centroid is also selected at random
from the dataset, the remaining centroids are sampled from the dataset
with a probability proportional to the squared distance from the closest
centroids. More precisely, the probability of selecting a point $x$
as the next centroid $c_{a}$ when $a\ge2$ is $\mathbb{P}\left(c_{a}=x\vert\left(c_{b}\right)_{b<a}\right)\propto\min_{b<a}\left\Vert x-c_{b}\right\Vert ^{2}$.
This procedure thus also computes the optimal partition as a byproduct,
like \noun{maxmin}. It can be further extended in a greedy manner:
at each step $a\ge2$, $s$ candidates are sampled according to the
previous probability distribution, the new $\mathrm{SSE}$ (with $a$
clusters) is computed, and the candidate with the lowest $\mathrm{SSE}$
is chosen. The number of candidates per step $s$ is usually logarithmic
in the number of clusters $k$; in all our tests, we have used $s=\left\lfloor 2+\log k\right\rfloor $.\footnote{This is the default value used by the scikit-learn library \citep{scikit-learn}
and seems to work well; \citet{celebi2013comparative} used
$s=\log\left(k\right)$ (it's unclear if it was truncated or rounded).} We refer to the original variant as \noun{km++} and to the greedy
one as \noun{gkm++}.

The computational complexity of \noun{km++} is
$O\left(kND\right)$ and it requires $1$ NDC, like \noun{maxmin},
while \noun{gkm++} scales like $O\left(kND\log k\right)$ and the
NDC required are slightly less than the number of candidates $s$.
In \citet{celebi2013comparative} \noun{gkm++} was reported
as superior to \noun{km++} and overall as one among the best linear
(in $N$) stochastic methods; in \citet{franti2019much} the
results of \noun{km++} were considered comparable to or slightly worse
than \noun{maxmin}; however, \noun{gkm++} was not tested.

\bfnoun{ref(init).} Bradley and Fayyad's ``refine'' seeding
algorithm~\citep{bradley1998refining} is the one that most resembles
our proposed scheme. Indeed, the initial step of the two methods is
basically the same, i.e. it consists in splitting the dataset into
$J$ random subsets and clustering them individually with $k$-means,
thus obtaining $J$ groups of $k$ centroids. The crucial difference
relies in the way in which these $J$ solutions are merged, which
in \citet{bradley1998refining} is referred to as a ``smoothing''
procedure: in the refine method, the whole pool of $Jk$ centroids
is used as a new dataset and clustered for $J$ times. Each time,
one of the previous centroid configurations is used as seed for the
Lloyd algorithm. Out of the resulting $J$ configurations, the one
with the smallest cost (computed on the pooled dataset) is finally
chosen. Originally, the authors used \noun{unif} as the seeding method
for clustering the subsets, but this can be trivially generalized
to other methods. We thus consider it a meta-method like PNN-smoothing,
and denote it with \noun{ref(init)} where \noun{init} is the seeding
method for the initial step.

If the computational cost of \noun{init}
is linear, $O\left(kND\right)$, then \noun{ref(init)} scales as $O\left(\left(kN+Jk^{2}\right)D\right)$.
The algorithm requires $J\le N/k$ in order to perform the first step,
and thus \noun{ref(init)} is always at most $O\left(kND\right)$.
In terms of NDC, the initial $J$ clusterings overall require at least
the same amount as \noun{init} (e.g. $1$ if \noun{init} is \noun{unif}
or \noun{maxmin}) plus some additional ones for the Lloyd's iteration
which are hard to estimate; then at least another $Jk/N$ NDC are
needed for the merging; finally, $1$ NDC is required at the end;
overall, the lower bound on the NDC is $1+Jk/N$ more than \noun{init}.
In the original publication, \citet{bradley1998refining}, only
\noun{ref(unif)} with $J=10$ was tested. This was also the setting
used in \citet{celebi2013comparative,franti2019much}. We also
used the same value of $J$ in our tests (the only exception being
a dataset for which $N/k=10$, in which case we used $J=5$), but
we tested more \noun{init} algorithms. As mentioned above, in \citet{celebi2013comparative}
\noun{ref(unif)} was found to be among the best methods, whereas in
\citet{franti2019much} it was shown to perform rather poorly
on synthetic datasets.

\bfnoun{pnn.} Our method starts out identically to \noun{ref(init)},
but it uses the PNN procedure to merge the resulting $J$ clusterings.
This procedure was originally introduced in \citet{equitz1989new}
as a seeding algorithm. The original algorithm, which we call \noun{pnn},
is hierarchical. It starts with $N$ clusters, one cluster per data
point, and then merges pairs of clusters iteratively until only $k$
clusters remain. Merging two clusters means that the partition $\mathcal{P}$
is updated by substituting the two clusters with their union. The
centroids set $\mathcal{C}$ is also updated, by substituting the
two starting centroids $c_{a}$ and $c_{b}$ with the centroid of
the new cluster, $c_{\mathrm{new}}=\left(z_{a}c_{a}+z_{b}c_{b}\right)/\left(z_{a}+z_{b}\right)$,
where $z_{a}$ and $z_{b}$ are the number of elements in each of
the two original clusters. The algorithm is deterministic and greedy:
the two clusters to be merged at each step are those whose merging
will result in the smallest increase in the $\mathrm{SSE}$ cost.
It is easy to see from eq.~\ref{eq:SSE} that the cost increment
of merging two clusters of sizes $z_{a}$ and $z_{b}$ and with centroids
$c_{a}$ and $c_{b}$ is $\Delta_{ab}=z_{a}z_{b}\left\Vert c_{a}-c_{b}\right\Vert ^{2}/\left(z_{a}+z_{b}\right)$.
Thus both the pairwise merging costs and the new centroid can be computed
using only the centroids and the cluster sizes.

The initial computation
of all the merging costs $\Delta_{ab}$ requires $N\left(N-1\right)/2$
distance computations, thus $O\left(DN^{2}\right)$ operations. The
computational complexity of a merging step, assuming that we are going
from $\hat{k}+1$ clusters to $\hat{k}$ clusters, would be $O\left(D\hat{k}^{2}\right)$
if performed straightforwardly, due to the need to update the $\Delta_{ab}$
after each merge. However, in \citet{franti1998fast} it was
shown that a significant speedup can be obtained by considering that
most cluster pairs are unaffected by individual merges, and the complexity
can be reduced to $O\left(D\hat{k}\tau_{\hat{k}}\right)$ where $\tau_{\hat{k}}\in\left[1,\hat{k}\right]$
is essentially the number of (potentially) affected clusters and is
generally much smaller than $\hat{k}$. The number of distance computations
of a merging step can easily be limited to exactly $\hat{k}$ by memoizing
and keeping up-to-date all the distances; however, since $\tau_{\hat{k}}$
is usually very small, in practice this technique only helps the running
time to a limited extent, and only in very high-dimensional cases,
and thus for simplicity our implementation does not use it.\footnote{Significantly more distance computations could be skipped by keeping
lower and upper bounds based on the triangle inequality, similarly
to the approach used to accelerate Lloyd's iterations in \citet{elkan2003using}
and others. This appears to be a promising optimization for high-dimensional
data, which is left for future work. } The update operations can be straightforwardly parallelized over
$\hat{k}$, which proves advantageous above a certain threshold. In
the original \noun{pnn} algorithm the merging step must be performed
$N-k$ times with $\hat{k}$ ranging from $N-1$ to $k$, which amounts
at roughly $N^{2}/2$ distance computations (assuming $N\gg k$).
Thus, the overall computational complexity is $\Omega\left(DN^{2}\right)$
and the NDC required are about $N/k$. This is quite expensive for
large datasets. On the other hand, the results in terms of the $\mathrm{SSE}$
objective are generally very good.

The \noun{PNN} scheme was also employed in the genetic algorithm of
\citet{franti2000genetic}, where however it was used as a crossover
step rather than a seeding procedure. In that algorithm, two given
configurations with $k$ clusters each are first merged into a single
configuration with $2k$ clusters, which is then used as the starting
point for the PNN iterative merging, until $k$ clusters remain. The
resulting cost is $O\left(Dk^{2}\tau_{k}\right)$, which (crucially)
does not involve a factor of $N$ thanks to the fact that only centroid
computations are involved in the merge. Our seeding scheme is similar
to this, in that we also use the iterative PNN merging procedure with
an initial number of clusters much smaller than $N$.

\begin{table}
\caption{Summary of seeding methods characteristics:
computational complexity (only $N$ and $k$ dependence, all methods
are linear in $D$) and normalized distance computations. For \noun{ref(init)}
and \noun{pnns(init)} only lower bounds on NDCs are available since
the expressions don't account for the internal Lloyd's iterations.
The $\tau_{X}$ terms for \noun{pnns(init)} and \noun{pnn} represent
hard-to-estimate terms, upper-bounded by $X$ but much smaller in
practice. For \noun{gkm++} we set $s=\left\lfloor 2+\log k\right\rfloor $.
For \noun{ref(init)} we normally set $J=10$. For \noun{pnns(init)}
we used the value $\rho=1$ throughout the main text.}
\label{tab:summary-of-seeding}
\begin{center}
\begin{tabular}{|c|c|c|}
\hline 
method & complexity & NDCs\tabularnewline
\hline 
\hline 
\noun{unif} & $kN$ & $1$\tabularnewline
\noun{maxmin} & $kN$ & $1$\tabularnewline
\noun{km++} & $kN$ & $1$\tabularnewline
\noun{gkm++} & $kNs$ & $1/k+s\left(k-1\right)/k$\tabularnewline
\noun{ref(init)} & $kN$ & $\ge\mathrm{NDC}\left(\text{\textsc{init}}\right)+1+Jk/N$\tabularnewline
\noun{pnns(init)} & $kN\tau_{\sqrt{kN}}$ & $\ge\mathrm{NDC}\left(\text{\textsc{init}}\right)+1+\rho/2$\tabularnewline
\noun{pnn} & $N^{2}\tau_{N}$ & $N/k$\tabularnewline
\hline 
\end{tabular}
\end{center}
\end{table}

\section{The PNN-smoothing scheme\label{sec:pnns}}

In this section we describe in detail the \noun{pnns(init)} seeding
scheme and discuss its properties.

\begin{figure}
\centering{}\includegraphics[width=1.0\textwidth]{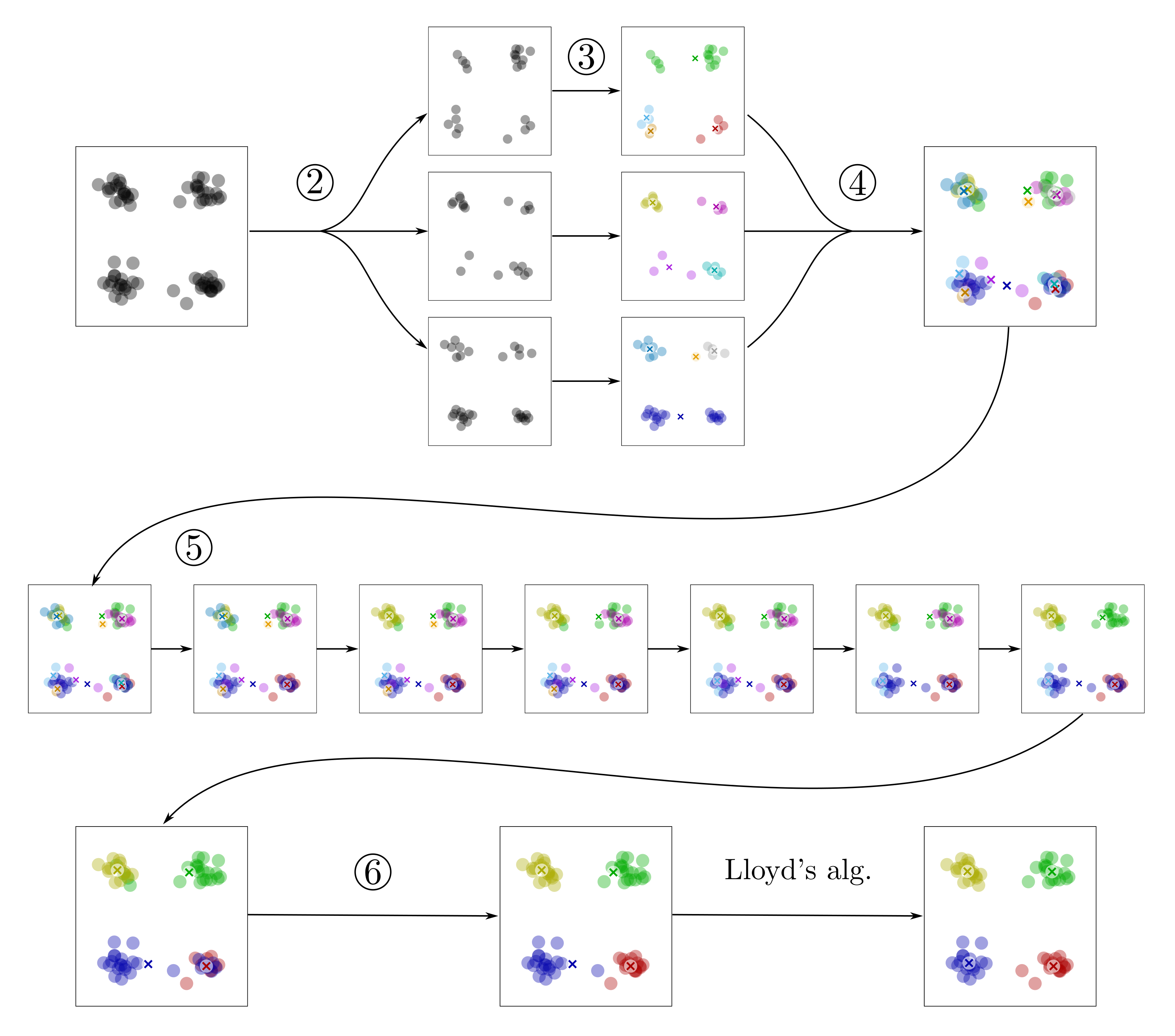}\caption{\label{fig:diagram}Example of \noun{pnns(init)} in action. Here $D=2$,
$N=72$, $k=4$ and \noun{init=unif}. The numbering of the steps follows
algorithm~\ref{alg:pnns} (step 1 simply yields $J=3$). The crosses represent centroids.
After step 3 the subsets clusterings contain some clear mistakes,
but after the merge (4) and the PNN procedure (5) the remaining 4
centroids are close to the correct positions. After step 6 the seeding
procedure is completed; the final local optimization solves the problem.}
\end{figure}

The inputs of the procedure are the same as for any other algorithm
(the dataset $\mathcal{X}$ and the number of clusters $k$), plus
the subset-seeding algorithm \noun{init}, and one extra parameter
$\rho$, used to determine the number of subsets $J$. The output
is a configuration to be used as a starting point for local optimization.
The high-level summary (see also the illustration of each step in
fig.~\ref{fig:diagram}) is as follows:

\begin{algorithm}[H]
\caption{PNNS seeding\label{alg:pnns}}
\DontPrintSemicolon

\KwIn{$\mathcal{X}$, $k$, \noun{init}, $\rho$}

%
%
%
%
%

\begin{enumerate}
\item Set $J=\left\lceil \sqrt{\rho N/\left(2k\right)}\right\rceil .$ Cap
the result at $\left\lfloor N/k\right\rfloor $.
\item Split $\mathcal{X}$ into $J$ random subsets $\left(\tilde{\mathcal{X}}_{a}\right)_{a=1..J}$.
\item Cluster each $\tilde{\mathcal{X}}_{a}$ independently, using \noun{init}
for seeding followed by Lloyd's algorithm; obtain $J$ configurations
$\left(\tilde{\mathcal{C}}_{a},\tilde{\mathcal{P}}_{a}\right)$, with
$k$ clusters each.
\item Collect the $J$ configurations $\left(\tilde{\mathcal{C}}_{a},\tilde{\mathcal{P}}_{a}\right)$
into a single configuration $\left(\mathcal{C}_{0},\mathcal{P}_{0}\right)$
for the entire $\mathcal{X}$, with $kJ$ centroids and clusters.
\item Merge the clusters of $\left(\mathcal{C}_{0},\mathcal{P}_{0}\right)$,
two at a time, using the PNN procedure, until $k$ clusters remain;
obtain a set of $k$ centroids $\mathcal{C}$.
\item Compute the optimal partition $\mathcal{P}$ associated to $\mathcal{C}$.
\end{enumerate}

\KwOut{$\left(\mathcal{C},\mathcal{P}\right)$}

\end{algorithm}


Next, we describe and discuss in more detail each step.
\begin{enumerate}
\item The parameterization in terms of $\rho$ rather than $J$ ensures
that the asymptotic behavior of the algorithm is almost linear in
$N$ and $k$, as discussed in the introduction (cf.~point 5 below).
The upper bound to $J$ is necessary for the following steps 2 and
3. The effective range of $\rho$, assuming for simplicity that $N$
is divisible by $k$, is $\left[2k/N,2N/k\right]$. The lower bound
leads to $J=1$ and \noun{pnns(init)=init}. The upper bound corresponds
to $J=N/k$ and \noun{pnns(init)=pnn} (under the assumption that \noun{init}
invoked on $k$ points will assign each one to its own cluster). In
other words, by changing $\rho$ we can interpolate between any seeding
algorithm \noun{init} and the pairwise-nearest-neighbor algorithm.
This explains why, as a rule of thumb, increasing $\rho$ improves
quality at the cost of performance, although this is not strictly
true in all cases.\\
The allowed range for $\rho$ is quite wide under normal circumstances,
in which $k$ is much smaller than $N$. Throughout the main text
we use the value $\rho=1$, which from our preliminary analysis seems
to result in a good trade-off in all cases and for all \noun{init}
algorithms. This is a valid (i.e.~non-degenerate) choice whenever
$k<N/2$, which is arguably always the case in realistic scenarios,
thus making it a reasonable default value. Additional results with
$\rho=10$ are reported in Appendix~\ref{Asec:tables}.
\item We split the data as evenly as possible, i.e.~we create $N\,\mod\,J$
subsets of size $\left\lfloor N/J\right\rfloor +1$ and $J-N\,\mod\,J$
subsets of size $\left\lfloor N/J\right\rfloor $. This is easy to
implement efficiently by just constructing a sorted list of indices,
each index being repeated for the appropriate number of times, and
shuffling it. Our preliminary testing showed that the algorithm is
not sensitive to the details of the splitting procedure.
\item The computational cost of each individual clustering of one of the
subsets depends on the choice of \noun{init}. Assuming that to be
linear, like in the examples mentioned in the previous section, this
scales like $N/J$ rather than $N$. Just like for refine, the NDC
required for this step are at least the same as for \noun{init} (thus
at least $1$) plus those for Lloyd's iterations, which are hard to
estimate. This step is also trivially parallelizable.
\item In order to obtain the new configuration, we just take the union of
the centroids, i.e. $\mathcal{C}_{0}=\bigcup_{a}\tilde{\mathcal{C}}_{a}$;
the partition $\mathcal{P}_{0}$ would also be simply the union of
the $\left(\tilde{\mathcal{P}}_{a}\right)_{a}$ with remapped indices,
but since it is not even needed for the algorithm (only the cluster
sizes are used) it can be skipped. It is also interesting to note
that the new configuration will, in general, be nowhere near optimal
for a problem with $kJ$ clusters, since points near each other will
likely be assigned to clusters coming from different subsets. Nevertheless,
at least under some favorable scenarios, we can expect that the centroids
in $\mathcal{C}_{0}$ may themselves be approximately clustered into
$k$ groups (see fig.~\ref{fig:diagram}).  This is the same intuition
at the root of the refine method. The question then becomes how to
best find a consensus configuration among the $J$ different results.
\item The PNN procedure will start, as the name implies, by merging the
closest centroids (accounting for their associated cluster size).
If the centroids in $\mathcal{C}_{0}$ are mostly clustered already,
the procedure will likely pick first the centroids that appeared in
multiple subset clusterings. Each time two centroids are merged, their
associated size (and thus weight) increases, such that even if in
the last stages some very sub-optimal partitions are merged with a
large cluster, the centroid will be heavily skewed toward the latter.\\
The computational cost of the merging scales like $O\left(D\left(Jk\right)^{2}\tau_{Jk}\right)=O\left(DNk\tau_{Jk}\right)$,
as per the analysis of the PNN procedure of the previous section.
This is indeed a consequence of our choice for the scaling of $J$.
The additional factor $\tau_{Jk}$ is hard to estimate; it is bounded
by $Jk=\sqrt{\rho Nk/2}$ but in practice it appears to be quite small.
This is the step that dominates the computational complexity of the
whole algorithm; however, it is seldom the step that takes up the
majority of computing time in practice when using $\rho=1$. Indeed,
the number of NDC can be estimated as just $\rho/2$ (where $\rho/4$
come from the initial step and $\rho/4$ from the merging process,
analogously to the analysis presented for \noun{pnn}).
\item At the end of the procedure the partition is recomputed ($1$ NDC).
This is done to compute the SSE and provide a consistent starting
point for Lloyd's algorithm (as discussed in the introduction), but
it also means that outlier points in the partitions that could be
(virtually) produced during the PNN merge are eliminated (like the
blue points in the bottom-right cluster and the green points in the
top-left cluster in fig.~\ref{fig:diagram}).
\end{enumerate}
Overall, the number of NDC in the \noun{pnns(init)} procedure is larger
by at least $1+\rho/2$ compared to \noun{init} (roughly, see steps
3, 5 and 6). This estimate appears to capture the biggest time penalty
of \noun{pnns(init)} in practice, even though it does not account
for the additional Lloyd's iterations performed in step 3. This is
because those usually require much less than $1$ NDC per iteration
thanks to the use of accelerator algorithms; furthermore they are
compensated by obtaining a seed which is closer to a local optimum,
so that the overall number of Lloyd's iterations at the end of the
process is comparable (the data showing this is reported in Appendix~\ref{Asec:tables}).

\section{Numerical experiments\label{sec:experiments}}

\subsection{Experimental setup\label{subsec:setup}}

We performed a series of tests comparing \noun{pnns} with all the
algorithms mentioned in sec.~\ref{sec:prior}, namely: \noun{unif},
\noun{maxmin}, \noun{km++}, \noun{gkm++}, \noun{pnn}, and \noun{ref(init)}
and \noun{pnns(init)} with \noun{init$\in\left\{ \textrm{\textsc{unif}},\textrm{\textsc{maxmin}},\textrm{\textsc{km++}},\textrm{\textsc{gkm++}}\right\} $}.
We used the same data structures and programming language (Julia v1.7.3)
for all of them. For the local optimization part, i.e. Lloyd's algorithm,
we implemented a number of techniques that can accelerate the computation,
while keeping it exact, by skipping some updates at the cost of some
bookkeeping: the ``reduced computation'' method (\noun{rc}) by \citet{kaukoranta1999reduced};
the Elkan method (\noun{elk}) from \citet{elkan2003using};
the Hamerly method (\noun{ham}) from \citet{hamerly2010making};
the Yinyang method (\noun{yy}) from \citet{ding2015yinyang};
the exponion method (\noun{exp}) from \citet{newling2016fast};
the ball-kmeans method (\noun{ball}) from \citet{xia2020fast}.
For \noun{elk} and \noun{yy} we implemented the simplified versions
described in \citet{newling2016fast}. For each of the datasets
that we tested we chose the accelerator technique that resulted in
the fastest average convergence time, when using \noun{km++} for seeding,
across at least $30$ random repetitions, and used that accelerator
for all other tests with that dataset. The winner for each dataset
is reported in Table~\ref{tab:datasets}. This choice puts \noun{pnns}
at the maximum disadvantage, since, as a general tendency, the better
the accelerator method, the larger the relative penalty of applying
the \noun{pnns} scheme.\footnote{This is because, as mentioned above, the number of overall Lloyd's
iteration ends up being comparable between \noun{init} and \noun{pnns(init)},
but accelerators become more effective as the iterations progress,
and in \noun{pnns} Lloyd's scheme is effectively started twice (this
effect is indeed captured by the NDC count discussed in sec.~\ref{sec:pnns},
point 6).}  In all our tests Lloyd's algorithm was run until convergence to
a fixed point. Our code is available at \href{https://github.com/carlobaldassi/KMeansPNNSmoothing.jl}{https://github.com/carlobaldassi/KMeansPNNSmoothing.jl}.

All the timings that we report refer to tests performed on the same
hardware (Intel Core i7-9750H 2.60GHz CPU with 6 physical cores, 64Gb
DDR4 2666MHz RAM, running Ubuntu Linux 20.04 with 5.15.0 kernel) with
no other computationally intensive processes running while testing.
All codes were carefully optimized\footnote{Our implementations of \noun{elk}, \noun{ham}, \noun{yy} and \noun{exp}
are generally roughly comparable to the very optimized ones (written
in C++) that accompany \citet{newling2016fast}, available at
\href{https://github.com/idiap/eakmeans}{https://github.com/idiap/eakmeans};
our implementation of \noun{ball} is generally faster than the C++
one provided by the original authors of \citet{xia2020fast},
available at \href{https://github.com/syxiaa/ball-k-means}{https://github.com/syxiaa/ball-k-means}.} and can run in parallel with multi-threading (the parallelization
is over the $N$ data points during Lloyd's iterations, and over the
$\hat{k}$ centroids when $\hat{k}\ge500$ during the PNN procedure,
see sec.~(\ref{sec:prior})); most of our results are shown for the
single-threaded case except where otherwise noted. We also report
the average total number of normalized distance computations (NDC),
summing up those performed during seeding (discussed in the previous
sections) and in the final optimization phase.

\begin{table}
\caption{Characteristics of the datasets used in the tests
and best accelerator for each dataset.}
\label{tab:datasets}
\begin{center}
\begin{tabular}{|cc|ccc|c|}
\hline 
\multicolumn{2}{|c|}{dataset} & $D$ & $N$ & $k$ & accel.\tabularnewline
\hline 
\hline 
\multirow{5}{*}{\begin{turn}{90}
synthetic
\end{turn}} & \emph{A3} & $2$ & $7500$ & $50$ & \noun{exp}\tabularnewline
 & \emph{Birch1} & $2$ & $100000$ & $100$ & \noun{exp}\tabularnewline
 & \emph{Birch2} & $2$ & $100000$ & $100$ & \noun{exp}\tabularnewline
 & \emph{Unbalance} & $2$ & $6500$ & $8$ & \noun{exp}\tabularnewline
 & \emph{Dim1024} & $1024$ & $1024$ & $16$ & \noun{exp}\tabularnewline
\hline 
\multirow{7}{*}{\begin{turn}{90}
real-world
\end{turn}} & \emph{Bridge} & 16 & $4096$ & $256$ & \noun{rc}\tabularnewline
 & \emph{House} & $3$ & $34112$ & $256$ & \noun{exp}\tabularnewline
 & \emph{Miss America} & $16$ & $6480$ & $256$ & \noun{rc}\tabularnewline
 & \emph{UrbanGB} & $2$ & $360177$ & $469$ & \noun{exp}\tabularnewline
 & \emph{Olivetti} & $4096$ & $400$ & $40$ & \noun{elk}\tabularnewline
 & \emph{Isolet} & $617$ & $7792$ & $26$ & \noun{elk}\tabularnewline
 & \emph{USCensus} & $68$ & $2458285$ & $100$ & \noun{yy}\tabularnewline
\hline 
\end{tabular}
\end{center}
\end{table}

We tested a number of synthetic and real-world datasets whose characteristics
are shown in table~\ref{tab:datasets}. The synthetic datasets are
mainly intended to measure the ability of the algorithms to find the
solution when one can be clearly identified, and for direct comparison
with the results of ref.~\citep{franti2019much}. In these kind of
datasets, when $k$-means gets stuck in a sub-optimal minimum, it
is usually due to having made one or more clearly identifiable mistakes
(see e.g. the supplementary materials of ref.~\citep{baldassi2022recombinator}).
As we shall show in sec.~\ref{subsec:synthetic}, \noun{pnns(init)}
is the only family of algorithms (among the linear or quasi-linear
ones) that is capable of finding the solution in 100\% of these cases;
this gives a degree of confidence that, when using \noun{pnns}, $k$-means
will not get stuck into clear and easily-fixed mistakes, and that
any variation in the optimization result is more likely to emerge
from the features of the datasets not being aligned with the SSE objective.

The real-world datasets on the other hand do not generally have a
simple structure or a simple ``solution'', and their optimum SSE
is unknown. Our tests in sec.~\ref{subsec:real-world} will show
that the \noun{pnns} scheme can provide systematic improvements in
the SSE with a bounded increase in computing time, generally offering
a better trade-off than alternative methods.

\subsection{Synthetic datasets\label{subsec:synthetic}}

\begin{table}
\caption{Results on synthetic datasets}
\label{tab:results1}
\begin{center}
\begin{tabular}{|c|lllll|}
\multicolumn{1}{c}{} & \multicolumn{5}{l}{success rate}\tabularnewline
\cline{2-6} \cline{3-6} \cline{4-6} \cline{5-6} \cline{6-6} 
\multicolumn{1}{c|}{} & \emph{A3} & \emph{Birch1} & \emph{Birch2} & \emph{Unbalance} & \emph{Dim1024}\tabularnewline
\hline 
\noun{\footnotesize{}unif} & {\footnotesize{}$0$} & {\footnotesize{}$0$} & {\footnotesize{}$0$} & {\footnotesize{}$0$} & {\footnotesize{}$0.001$}\tabularnewline
\noun{\footnotesize{}maxmin} & {\footnotesize{}$0.004$} & {\footnotesize{}$0$} & {\footnotesize{}$0$} & {\footnotesize{}$0.223$} & {\footnotesize{}$1$}\tabularnewline
\noun{\footnotesize{}km++} & {\footnotesize{}$0$} & {\footnotesize{}$0$} & {\footnotesize{}$0$} & {\footnotesize{}$0.541$} & {\footnotesize{}$0.996$}\tabularnewline
\noun{\footnotesize{}gkm++} & {\footnotesize{}$0.051$} & {\footnotesize{}$0.02$} & {\footnotesize{}$0.06$} & {\footnotesize{}$0.946$} & {\footnotesize{}$1$}\tabularnewline
\noun{\footnotesize{}ref(unif)} & {\footnotesize{}$0$} & {\footnotesize{}$0$} & {\footnotesize{}$0$} & {\footnotesize{}$0$} & {\footnotesize{}$0.055$}\tabularnewline
\noun{\footnotesize{}ref(gkm++)} & {\footnotesize{}$0.239$} & {\footnotesize{}$0.03$} & {\footnotesize{}$0.44$} & {\footnotesize{}$1$} & {\footnotesize{}$1$}\tabularnewline
\noun{\footnotesize{}pnns(unif)} & {\footnotesize{}$0.709$} & {\footnotesize{}$1$} & {\footnotesize{}$0.26$} & {\footnotesize{}$0.696$} & {\footnotesize{}$0.919$}\tabularnewline
\noun{\footnotesize{}pnns(maxmin)} & {\footnotesize{}$1$} & {\footnotesize{}$1$} & {\footnotesize{}$1$} & {\footnotesize{}$1$} & {\footnotesize{}$1$}\tabularnewline
\noun{\footnotesize{}pnns(km++)} & {\footnotesize{}$0.98$} & {\footnotesize{}$1$} & {\footnotesize{}$1$} & {\footnotesize{}$1$} & {\footnotesize{}$1$}\tabularnewline
\noun{\footnotesize{}pnns(gkm++)} & {\footnotesize{}$1$} & {\footnotesize{}$1$} & {\footnotesize{}$1$} & {\footnotesize{}$1$} & {\footnotesize{}$1$}\tabularnewline
\noun{\footnotesize{}pnn} & {\footnotesize{}$1$} & {\footnotesize{}$1$} & {\footnotesize{}$1$} & {\footnotesize{}$1$} & {\footnotesize{}$1$}\tabularnewline
\hline 
\multicolumn{1}{c}{} & \multicolumn{5}{l}{convergence time {\scriptsize{}(mean$\pm$stdev)\vphantom{{\huge{}|}}}}\tabularnewline
\cline{2-6} \cline{3-6} \cline{4-6} \cline{5-6} \cline{6-6} 
\multicolumn{1}{c|}{} & \emph{A3} & \emph{Birch1} & \emph{Birch2} & \emph{Unbalance} & \emph{Dim1024}\tabularnewline
\multicolumn{1}{c|}{} & {\scriptsize{}(in $10^{-3}s$)} & {\scriptsize{}(in $10^{-1}s$)} & {\scriptsize{}(in $10^{-1}s$)} & {\scriptsize{}(in $10^{-3}s$)} & {\scriptsize{}(in $10^{-3}s$)}\tabularnewline
\hline 
\noun{\footnotesize{}unif} & {\footnotesize{}$5.1\pm1.5$} & {\footnotesize{}$2.3\pm0.6$} & {\footnotesize{}$0.63\pm0.10$} & {\footnotesize{}$3.1\pm1.8$} & {\footnotesize{}$5.5\pm1.3$}\tabularnewline
\noun{\footnotesize{}maxmin} & {\footnotesize{}$4.0\pm0.9$} & {\footnotesize{}$2.0\pm0.4$} & {\footnotesize{}$0.75\pm0.10$} & {\footnotesize{}$0.7\pm0.8$} & {\footnotesize{}$5.4\pm0.8$}\tabularnewline
\noun{\footnotesize{}km++} & {\footnotesize{}$4.1\pm1.2$} & {\footnotesize{}$1.78\pm0.38$} & {\footnotesize{}$0.66\pm0.10$} & {\footnotesize{}$1.1\pm1.3$} & {\footnotesize{}$5.4\pm1.5$}\tabularnewline
\noun{\footnotesize{}gkm++} & {\footnotesize{}$6.5\pm1.4$} & {\footnotesize{}$2.44\pm0.26$} & {\footnotesize{}$1.64\pm0.11$} & {\footnotesize{}$0.9\pm1.0$} & {\footnotesize{}$12.8\pm1.7$}\tabularnewline
\noun{\footnotesize{}ref(unif)} & {\footnotesize{}$14.7\pm2.5$} & {\footnotesize{}$2.97\pm0.33$} & {\footnotesize{}$1.39\pm0.09$} & {\footnotesize{}$4.7\pm1.5$} & {\footnotesize{}$20.5\pm3.4$}\tabularnewline
\noun{\footnotesize{}ref(gkm++)} & {\footnotesize{}$13.8\pm2.2$} & {\footnotesize{}$3.06\pm0.23$} & {\footnotesize{}$2.06\pm0.12$} & {\footnotesize{}$2.0\pm1.3$} & {\footnotesize{}$19.1\pm3.0$}\tabularnewline
\noun{\footnotesize{}pnns(unif)} & {\footnotesize{}$11.2\pm2.0$} & {\footnotesize{}$2.79\pm0.09$} & {\footnotesize{}$1.78\pm0.11$} & {\footnotesize{}$3.7\pm1.6$} & {\footnotesize{}$14.2\pm2.9$}\tabularnewline
\noun{\footnotesize{}pnns(maxmin)} & {\footnotesize{}$8.4\pm1.7$} & {\footnotesize{}$2.51\pm0.10$} & {\footnotesize{}$1.60\pm0.09$} & {\footnotesize{}$1.8\pm1.4$} & {\footnotesize{}$12.9\pm2.5$}\tabularnewline
\noun{\footnotesize{}pnns(km++)} & {\footnotesize{}$9.1\pm1.8$} & {\footnotesize{}$2.55\pm0.07$} & {\footnotesize{}$1.59\pm0.07$} & {\footnotesize{}$2.0\pm1.2$} & {\footnotesize{}$13.1\pm2.8$}\tabularnewline
\noun{\footnotesize{}pnns(gkm++)} & {\footnotesize{}$12.1\pm2.0$} & {\footnotesize{}$3.32\pm0.08$} & {\footnotesize{}$2.49\pm0.09$} & {\footnotesize{}$2.5\pm1.4$} & {\footnotesize{}$19.9\pm3.3$}\tabularnewline
\noun{\footnotesize{}pnn} & {\footnotesize{}$396\pm16$} & {\footnotesize{}$660\pm6$} & {\footnotesize{}$709.6\pm3.1$} & {\footnotesize{}$292\pm11$} & {\footnotesize{}$1678\pm34$}\tabularnewline
\hline 
\multicolumn{1}{c}{} & \multicolumn{5}{l}{normalized distance computations {\scriptsize{}(mean$\pm$stdev)\vphantom{{\huge{}|}}}}\tabularnewline
\cline{2-6} \cline{3-6} \cline{4-6} \cline{5-6} \cline{6-6} 
\multicolumn{1}{c|}{} & \emph{A3} & \emph{Birch1} & \emph{Birch2} & \emph{Unbalance} & \emph{Dim1024}\tabularnewline
\hline 
\noun{\footnotesize{}unif} & {\footnotesize{}$1.80\pm0.18$} & {\footnotesize{}$2.52\pm0.37$} & {\footnotesize{}$1.146\pm0.027$} & {\footnotesize{}$2.7\pm0.6$} & {\footnotesize{}$1.89\pm0.32$}\tabularnewline
\noun{\footnotesize{}maxmin} & {\footnotesize{}$1.33\pm0.09$} & {\footnotesize{}$2.08\pm0.29$} & {\footnotesize{}$1.073\pm0.020$} & {\footnotesize{}$1.26\pm0.10$} & {\footnotesize{}$1.07812\pm0.0$}\tabularnewline
\noun{\footnotesize{}km++} & {\footnotesize{}$1.41\pm0.12$} & {\footnotesize{}$2.00\pm0.25$} & {\footnotesize{}$1.063\pm0.014$} & {\footnotesize{}$1.35\pm0.31$} & {\footnotesize{}$1.079\pm0.009$}\tabularnewline
\noun{\footnotesize{}gkm++} & {\footnotesize{}$5.08\pm0.05$} & {\footnotesize{}$6.46\pm0.16$} & {\footnotesize{}$5.972\pm0.004$} & {\footnotesize{}$3.78\pm0.11$} & {\footnotesize{}$3.89062\pm0.0$}\tabularnewline
\noun{\footnotesize{}ref(unif)} & {\footnotesize{}$3.94\pm0.12$} & {\footnotesize{}$3.75\pm0.21$} & {\footnotesize{}$2.393\pm0.019$} & {\footnotesize{}$4.6\pm0.5$} & {\footnotesize{}$6.6\pm0.4$}\tabularnewline
\noun{\footnotesize{}ref(gkm++)} & {\footnotesize{}$6.992\pm0.029$} & {\footnotesize{}$7.72\pm0.11$} & {\footnotesize{}$7.1128\pm0.0030$} & {\footnotesize{}$5.068\pm0.028$} & {\footnotesize{}$6.9261\pm0.0014$}\tabularnewline
\noun{\footnotesize{}pnns(unif)} & {\footnotesize{}$3.84\pm0.06$} & {\footnotesize{}$4.07\pm0.04$} & {\footnotesize{}$3.191\pm0.012$} & {\footnotesize{}$4.63\pm0.37$} & {\footnotesize{}$4.27\pm0.16$}\tabularnewline
\noun{\footnotesize{}pnns(maxmin)} & {\footnotesize{}$3.283\pm0.025$} & {\footnotesize{}$3.662\pm0.025$} & {\footnotesize{}$3.044\pm0.007$} & {\footnotesize{}$3.257\pm0.026$} & {\footnotesize{}$3.380\pm0.027$}\tabularnewline
\noun{\footnotesize{}pnns(km++)} & {\footnotesize{}$3.439\pm0.039$} & {\footnotesize{}$3.791\pm0.031$} & {\footnotesize{}$3.084\pm0.008$} & {\footnotesize{}$3.40\pm0.05$} & {\footnotesize{}$3.380\pm0.027$}\tabularnewline
\noun{\footnotesize{}pnns(gkm++)} & {\footnotesize{}$7.137\pm0.022$} & {\footnotesize{}$8.416\pm0.020$} & {\footnotesize{}$7.957\pm0.007$} & {\footnotesize{}$5.903\pm0.034$} & {\footnotesize{}$6.193\pm0.027$}\tabularnewline
\noun{\footnotesize{}pnn} & {\footnotesize{}$262.324$} & {\footnotesize{}$1730.39$} & {\footnotesize{}$1738.25$} & {\footnotesize{}$1422.21$} & {\footnotesize{}$531.759$}\tabularnewline
\hline 
\end{tabular}
\end{center}
\end{table}

\begin{table*}[!tp]
\caption{Results on real-world datasets}
\label{tab:results2}
\begin{center}
\begin{tabular}{|c|llllll|}
\multicolumn{1}{c}{} & \multicolumn{6}{l}{mean $\mathrm{SSE}$ cost {\scriptsize{}(mean$\pm$stdev over $100$
repetitions)}}\tabularnewline
\cline{2-7} \cline{3-7} \cline{4-7} \cline{5-7} \cline{6-7} \cline{7-7} 
\multicolumn{1}{c|}{} & \emph{Bridge} & \emph{House} & \emph{Miss A.} & \emph{Urb.GB} & \emph{Olivetti} & \emph{Isolet}\tabularnewline
\multicolumn{1}{c|}{} & {\scriptsize{}($\times10^{7}$)} & {\scriptsize{}($\times10^{5}$)} & {\scriptsize{}($\times10^{5}$)} & {\scriptsize{}($\times10^{2}$)} & {\scriptsize{}($\times10^{4}$)} & {\scriptsize{}($\times10^{5}$)}\tabularnewline
\hline 
\noun{\footnotesize{}unif} & {\footnotesize{}$1.178\pm0.009$} & {\footnotesize{}$10.11\pm0.13$} & {\footnotesize{}$6.07\pm0.05$} & {\footnotesize{}$6.8\pm1.1$} & {\footnotesize{}$1.296\pm0.027$} & {\footnotesize{}$1.196\pm0.010$}\tabularnewline
\noun{\footnotesize{}maxmin} & {\footnotesize{}$1.139\pm0.005$} & {\footnotesize{}$10.18\pm0.07$} & {\footnotesize{}$5.794\pm0.028$} & {\footnotesize{}$2.99\pm0.07$} & {\footnotesize{}$1.252\pm0.015$} & {\footnotesize{}$1.233\pm0.014$}\tabularnewline
\noun{\footnotesize{}km++} & {\footnotesize{}$1.154\pm0.007$} & {\footnotesize{}$9.60\pm0.04$} & {\footnotesize{}$5.687\pm0.037$} & {\footnotesize{}$2.72\pm0.06$} & {\footnotesize{}$1.276\pm0.021$} & {\footnotesize{}$1.196\pm0.010$}\tabularnewline
\noun{\footnotesize{}gkm++} & {\footnotesize{}$1.124\pm0.004$} & {\footnotesize{}$9.529\pm0.027$} & {\footnotesize{}$5.507\pm0.016$} & {\footnotesize{}$2.430\pm0.020$} & {\footnotesize{}$1.227\pm0.013$} & {\footnotesize{}$1.190\pm0.007$}\tabularnewline
\noun{\footnotesize{}ref(unif)} & {\footnotesize{}$1.162\pm0.006$} & {\footnotesize{}$9.95\pm0.10$} & {\footnotesize{}$5.850\pm0.035$} & {\footnotesize{}$5.3\pm0.4$} & {\footnotesize{}$1.298\pm0.023$} & {\footnotesize{}$1.189\pm0.006$}\tabularnewline
\noun{\footnotesize{}ref(gkm++)} & {\footnotesize{}$1.159\pm0.005$} & {\footnotesize{}$9.562\pm0.027$} & {\footnotesize{}$5.826\pm0.031$} & {\footnotesize{}$2.420\pm0.022$} & {\footnotesize{}$1.255\pm0.020$} & {\footnotesize{}$1.1837\pm0.0036$}\tabularnewline
\noun{\footnotesize{}pnns(unif)} & {\footnotesize{}$1.124\pm0.005$} & {\footnotesize{}$9.553\pm0.034$} & {\footnotesize{}$5.607\pm0.040$} & {\footnotesize{}$2.70\pm0.06$} & {\footnotesize{}$1.215\pm0.014$} & {\footnotesize{}$1.1800\pm0.0019$}\tabularnewline
\noun{\footnotesize{}pnns(km++)} & {\footnotesize{}$1.1076\pm0.0033$} & {\footnotesize{}$9.486\pm0.020$} & {\footnotesize{}$5.403\pm0.013$} & {\footnotesize{}$2.323\pm0.010$} & {\footnotesize{}$1.209\pm0.013$} & {\footnotesize{}$1.1795\pm0.0016$}\tabularnewline
\noun{\footnotesize{}pnns(gkm++)} & {\footnotesize{}$1.0947\pm0.0029$} & {\footnotesize{}$9.476\pm0.020$} & {\footnotesize{}$5.342\pm0.010$} & {\footnotesize{}$2.297\pm0.004$} & {\footnotesize{}$1.189\pm0.007$} & {\footnotesize{}$1.1790\pm0.0015$}\tabularnewline
\noun{\footnotesize{}pnn} & {\footnotesize{}$1.08279$} & {\footnotesize{}$9.49701$} & {\footnotesize{}$5.31588$} & {\footnotesize{}$2.3153$} & {\footnotesize{}$1.16238$} & {\footnotesize{}$1.17692$}\tabularnewline
\hline 
\multicolumn{1}{c}{} & \multicolumn{6}{l}{minimum $\mathrm{SSE}$ cost {\scriptsize{}(over $100$ repetitions)\vphantom{{\huge{}|}}}}\tabularnewline
\cline{2-7} \cline{3-7} \cline{4-7} \cline{5-7} \cline{6-7} \cline{7-7} 
\multicolumn{1}{c|}{} & \emph{Bridge} & \emph{House} & \emph{Miss A.} & \emph{Urb.GB} & \emph{Olivetti} & \emph{Isolet}\tabularnewline
\multicolumn{1}{c|}{} & {\scriptsize{}($\times10^{7}$)} & {\scriptsize{}($\times10^{5}$)} & {\scriptsize{}($\times10^{5}$)} & {\scriptsize{}($\times10^{2}$)} & {\scriptsize{}($\times10^{4}$)} & {\scriptsize{}($\times10^{5}$)}\tabularnewline
\hline 
\noun{\footnotesize{}unif} & {\footnotesize{}$1.15723$} & {\footnotesize{}$9.87713$} & {\footnotesize{}$5.95536$} & {\footnotesize{}$5.42031$} & {\footnotesize{}$1.23747$} & {\footnotesize{}$1.1795$}\tabularnewline
\noun{\footnotesize{}maxmin} & {\footnotesize{}$1.12869$} & {\footnotesize{}$10.0159$} & {\footnotesize{}$5.72134$} & {\footnotesize{}$2.90322$} & {\footnotesize{}$1.21735$} & {\footnotesize{}$1.20639$}\tabularnewline
\noun{\footnotesize{}km++} & {\footnotesize{}$1.14219$} & {\footnotesize{}$9.52638$} & {\footnotesize{}$5.59949$} & {\footnotesize{}$2.60788$} & {\footnotesize{}$1.23035$} & {\footnotesize{}$1.18025$}\tabularnewline
\noun{\footnotesize{}gkm++} & {\footnotesize{}$1.11421$} & {\footnotesize{}$9.45992$} & {\footnotesize{}$5.47583$} & {\footnotesize{}$2.37869$} & {\footnotesize{}$1.19908$} & {\footnotesize{}$1.17709$}\tabularnewline
\noun{\footnotesize{}ref(unif)} & {\footnotesize{}$1.14562$} & {\footnotesize{}$9.77128$} & {\footnotesize{}$5.76604$} & {\footnotesize{}$4.39285$} & {\footnotesize{}$1.24007$} & {\footnotesize{}$1.17688$}\tabularnewline
\noun{\footnotesize{}ref(gkm++)} & {\footnotesize{}$1.14344$} & {\footnotesize{}$9.49776$} & {\footnotesize{}$5.74112$} & {\footnotesize{}$2.37821$} & {\footnotesize{}$1.2116$} & {\footnotesize{}$1.17737$}\tabularnewline
\noun{\footnotesize{}pnns(unif)} & {\footnotesize{}$1.11345$} & {\footnotesize{}$9.47461$} & {\footnotesize{}$5.51748$} & {\footnotesize{}$2.5921$} & {\footnotesize{}$1.17155$} & {\footnotesize{}$1.17685$}\tabularnewline
\noun{\footnotesize{}pnns(km++)} & {\footnotesize{}$1.10041$} & {\footnotesize{}$9.44324$} & {\footnotesize{}$5.37209$} & {\footnotesize{}$2.30731$} & {\footnotesize{}$1.18579$} & {\footnotesize{}$1.17687$}\tabularnewline
\noun{\footnotesize{}pnns(gkm++)} & {\footnotesize{}$1.08729$} & {\footnotesize{}$9.42885$} & {\footnotesize{}$5.31768$} & {\footnotesize{}$2.28822$} & {\footnotesize{}$1.17066$} & {\footnotesize{}$1.17695$}\tabularnewline
\noun{\footnotesize{}pnn} & {\footnotesize{}$1.08279$} & {\footnotesize{}$9.49701$} & {\footnotesize{}$5.31588$} & {\footnotesize{}$2.3153$} & {\footnotesize{}$1.16238$} & {\footnotesize{}$1.17692$}\tabularnewline
\hline 
\multicolumn{1}{c}{} & \multicolumn{6}{l}{convergence time {\scriptsize{}(mean$\pm$stdev)\vphantom{{\huge{}|}}}}\tabularnewline
\cline{2-7} \cline{3-7} \cline{4-7} \cline{5-7} \cline{6-7} \cline{7-7} 
\multicolumn{1}{c|}{} & \emph{Bridge} & \emph{House} & \emph{Miss A.} & \emph{Urb.GB} & \emph{Olivetti} & \emph{Isolet}\tabularnewline
\multicolumn{1}{c|}{} & {\scriptsize{}(in $10^{-2}s$)} & {\scriptsize{}(in $10^{-1}s$)} & {\scriptsize{}(in $10^{-1}s$)} & {\scriptsize{}(in $s$)} & {\scriptsize{}(in $10^{-2}s$)} & {\scriptsize{}(in $10^{-1}s$)}\tabularnewline
\hline 
\noun{\footnotesize{}unif} & {\footnotesize{}$4.17\pm0.38$} & {\footnotesize{}$3.3\pm0.6$} & {\footnotesize{}$1.13\pm0.09$} & {\footnotesize{}$1.63\pm0.23$} & {\footnotesize{}$2.7\pm0.6$} & {\footnotesize{}$1.52\pm0.30$}\tabularnewline
\noun{\footnotesize{}maxmin} & {\footnotesize{}$4.6\pm0.5$} & {\footnotesize{}$3.2\pm0.5$} & {\footnotesize{}$1.78\pm0.33$} & {\footnotesize{}$1.73\pm0.20$} & {\footnotesize{}$4.1\pm0.6$} & {\footnotesize{}$1.58\pm0.21$}\tabularnewline
\noun{\footnotesize{}km++} & {\footnotesize{}$4.22\pm0.36$} & {\footnotesize{}$2.40\pm0.37$} & {\footnotesize{}$1.15\pm0.11$} & {\footnotesize{}$1.56\pm0.13$} & {\footnotesize{}$4.10\pm0.39$} & {\footnotesize{}$1.75\pm0.29$}\tabularnewline
\noun{\footnotesize{}gkm++} & {\footnotesize{}$6.45\pm0.35$} & {\footnotesize{}$3.33\pm0.33$} & {\footnotesize{}$1.58\pm0.13$} & {\footnotesize{}$5.51\pm0.16$} & {\footnotesize{}$8.8\pm0.6$} & {\footnotesize{}$3.02\pm0.29$}\tabularnewline
\noun{\footnotesize{}ref(unif)} & {\footnotesize{}$22.0\pm0.5$} & {\footnotesize{}$6.5\pm0.4$} & {\footnotesize{}$3.26\pm0.11$} & {\footnotesize{}$3.65\pm0.28$} & {\footnotesize{}$8.0\pm0.5$} & {\footnotesize{}$1.68\pm0.24$}\tabularnewline
\noun{\footnotesize{}ref(gkm++)} & {\footnotesize{}$24.6\pm0.6$} & {\footnotesize{}$6.48\pm0.36$} & {\footnotesize{}$3.99\pm0.19$} & {\footnotesize{}$5.36\pm0.23$} & {\footnotesize{}$12.0\pm0.4$} & {\footnotesize{}$2.19\pm0.22$}\tabularnewline
\noun{\footnotesize{}pnns(unif)} & {\footnotesize{}$5.90\pm0.31$} & {\footnotesize{}$4.74\pm0.31$} & {\footnotesize{}$1.52\pm0.09$} & {\footnotesize{}$4.12\pm0.17$} & {\footnotesize{}$6.2\pm1.0$} & {\footnotesize{}$1.63\pm0.16$}\tabularnewline
\noun{\footnotesize{}pnns(km++)} & {\footnotesize{}$5.9\pm0.4$} & {\footnotesize{}$4.44\pm0.31$} & {\footnotesize{}$1.52\pm0.10$} & {\footnotesize{}$3.58\pm0.16$} & {\footnotesize{}$6.74\pm0.38$} & {\footnotesize{}$1.66\pm0.17$}\tabularnewline
\noun{\footnotesize{}pnns(gkm++)} & {\footnotesize{}$8.13\pm0.39$} & {\footnotesize{}$5.22\pm0.32$} & {\footnotesize{}$1.89\pm0.11$} & {\footnotesize{}$5.99\pm0.14$} & {\footnotesize{}$9.51\pm0.39$} & {\footnotesize{}$2.07\pm0.18$}\tabularnewline
\noun{\footnotesize{}pnn} & {\footnotesize{}$36.2\pm1.2$} & {\footnotesize{}$93.68\pm0.20$} & {\footnotesize{}$9.356\pm0.033$} & {\footnotesize{}$819\pm7$} & {\footnotesize{}$30.5\pm1.8$} & {\footnotesize{}$297.2\pm2.8$}\tabularnewline
\hline 
\multicolumn{1}{c}{} & \multicolumn{6}{l}{normalized distance computations {\scriptsize{}(mean$\pm$stdev)\vphantom{{\huge{}|}}}}\tabularnewline
\cline{2-7} \cline{3-7} \cline{4-7} \cline{5-7} \cline{6-7} \cline{7-7} 
\multicolumn{1}{c|}{} & \emph{Bridge} & \emph{House} & \emph{Miss A.} & \emph{Urb.GB} & \emph{Olivetti} & \emph{Isolet}\tabularnewline
\hline 
\noun{\footnotesize{}unif} & {\footnotesize{}$8.4\pm0.5$} & {\footnotesize{}$5.7\pm0.9$} & {\footnotesize{}$13.5\pm0.9$} & {\footnotesize{}$1.60\pm0.23$} & {\footnotesize{}$1.79\pm0.14$} & {\footnotesize{}$3.35\pm0.36$}\tabularnewline
\noun{\footnotesize{}maxmin} & {\footnotesize{}$8.9\pm0.8$} & {\footnotesize{}$4.5\pm0.6$} & {\footnotesize{}$23\pm4$} & {\footnotesize{}$1.196\pm0.037$} & {\footnotesize{}$2.21\pm0.07$} & {\footnotesize{}$3.14\pm0.25$}\tabularnewline
\noun{\footnotesize{}km++} & {\footnotesize{}$8.2\pm0.6$} & {\footnotesize{}$3.71\pm0.39$} & {\footnotesize{}$13.7\pm1.3$} & {\footnotesize{}$1.229\pm0.029$} & {\footnotesize{}$2.26\pm0.07$} & {\footnotesize{}$3.36\pm0.35$}\tabularnewline
\noun{\footnotesize{}gkm++} & {\footnotesize{}$13.2\pm0.5$} & {\footnotesize{}$9.27\pm0.32$} & {\footnotesize{}$19.5\pm1.5$} & {\footnotesize{}$8.168\pm0.033$} & {\footnotesize{}$6.04\pm0.05$} & {\footnotesize{}$6.92\pm0.33$}\tabularnewline
\noun{\footnotesize{}ref(unif)} & {\footnotesize{}$44.8\pm0.8$} & {\footnotesize{}$8.3\pm0.6$} & {\footnotesize{}$39.4\pm1.0$} & {\footnotesize{}$3.15\pm0.15$} & {\footnotesize{}$5.68\pm0.10$} & {\footnotesize{}$4.62\pm0.26$}\tabularnewline
\noun{\footnotesize{}ref(gkm++)} & {\footnotesize{}$49.5\pm0.8$} & {\footnotesize{}$12.25\pm0.35$} & {\footnotesize{}$48.3\pm2.1$} & {\footnotesize{}$9.465\pm0.021$} & {\footnotesize{}$10.34\pm0.07$} & {\footnotesize{}$8.20\pm0.22$}\tabularnewline
\noun{\footnotesize{}pnns(unif)} & {\footnotesize{}$10.81\pm0.37$} & {\footnotesize{}$6.80\pm0.36$} & {\footnotesize{}$16.6\pm0.9$} & {\footnotesize{}$4.03\pm0.09$} & {\footnotesize{}$4.25\pm0.10$} & {\footnotesize{}$5.06\pm0.15$}\tabularnewline
\noun{\footnotesize{}pnns(km++)} & {\footnotesize{}$10.6\pm0.4$} & {\footnotesize{}$6.05\pm0.29$} & {\footnotesize{}$16.7\pm1.0$} & {\footnotesize{}$3.31\pm0.06$} & {\footnotesize{}$5.00\pm0.10$} & {\footnotesize{}$5.24\pm0.17$}\tabularnewline
\noun{\footnotesize{}pnns(gkm++)} & {\footnotesize{}$15.9\pm0.4$} & {\footnotesize{}$11.59\pm0.30$} & {\footnotesize{}$22.2\pm1.2$} & {\footnotesize{}$10.191\pm0.014$} & {\footnotesize{}$8.76\pm0.09$} & {\footnotesize{}$8.73\pm0.16$}\tabularnewline
\noun{\footnotesize{}pnn} & {\footnotesize{}$51.8177$} & {\footnotesize{}$271.383$} & {\footnotesize{}$87.8464$} & {\footnotesize{}$1351$} & {\footnotesize{}$19.7431$} & {\footnotesize{}$739.388$}\tabularnewline
\hline 
\end{tabular}
\end{center}
\end{table*}

In \citet{franti2019much}, several synthetic datasets with
different characteristics were chosen and tested in order to probe
the strengths and weaknesses of several seeding algorithms with respect
to properties of the data. We picked 5 of the most challenging ones,
all obtained from the \href{http://cs.uef.fi/sipu/datasets/}{UEF repository}~\citep{franti2018k}:
\emph{A3}, \emph{Birch1},
\emph{Birch2}, \emph{Unbalance} and \emph{Dim1024} (see table~\ref{tab:datasets}).
We scaled each dataset uniformly in order to make them span the range
$\left[0,1\right]^{D}$. The difficulty for \emph{A3}, \emph{Birch1}
and \emph{Birch2} is in their relatively large size and abundance
of local minima; for \emph{Unbalance}, it's the fact that some clusters
are small and far from the bigger ones; for \emph{Dim1024} it's the
large dimensionality. All algorithms tested in~\citet{franti2019much}
showed poor results in at least some of these datasets; in particular,
the authors report a $0\%$ success rate (as defined below) on \emph{Birch1}
for all algorithms. 

All of these datasets were generated from isotropic Gaussians centered
around ground-truth centroids, and for all of them the global optimum
of the SSE is very close to the ground truth. Under these circumstances,
it is reasonable to classify the local minima configurations that
the algorithms produce by their ``centroid index'' (CI), as defined
in \citet{franti2014centroid}. The CI is computed by matching
each centroid of a clustering with its closest one from the ground
truth, and counting the number of unmatched ground truth centroids;
in formulas:
\global\long\def\ind{\mathbbm{1}}%
\begin{align}
\mathrm{CI} & =\sum_{b=1}^{k^{\mathrm{gt}}}\ind\left(\nexists a\in\left\{ 1,\dots,k\right\} \,:\,b=\argmin_{1\le b^{\prime}\le k^{\mathrm{gt}}}d\left(c_{a},c_{b^{\prime}}^{\mathrm{gt}}\right)^{2}\right)\label{eq:CI}
\end{align}
where $c_{b}^{\mathrm{gt}}$ with $b\in\left\{ 1,\dots,k^{\mathrm{gt}}\right\} $
are the ground-truth centroids (in our setup $k=k^{\mathrm{gt}}$),
and $\ind\left(\cdot\right)$ is an indicator function. The CI can
be interpreted as the number of mistakes in the resulting clustering,
and therefore we define the success rate of an algorithm as the frequency
with which it finds a solution, i.e. a configuration with $\mathrm{CI}=0$.

We present our most representative results in table~\ref{tab:results1};
the complete results are reported in Appendix~\ref{Asubsec:tables-synth}.
All algorithms except for \noun{pnn} and those in the \noun{pnns}
family fail badly in at least some dataset. Conversely, all the \noun{pnns(init)}
algorithms solve all the datasets in 100\% of the cases (98\% for
\noun{pnns(km++)} on \emph{A3}), with the only notable exception being
\noun{init=unif}. Even in that case, however, the performance is much
better than \noun{unif} and \noun{ref(unif)}, and at times even than
\noun{ref(gkm++)} (which is the variant in the refine family that
gives the best overall results). This demonstrates that (at least
for this scenario) PNN-smoothing is a considerably better smoothing
technique than refine.

The timings of the \noun{pnns(init)} algorithms are generally comparable
with, and often better than, those of the refine family, and within
a constant factor (smaller than $3$) of the basic methods; on the
other hand, the only other algorithm capable of solving all the datasets,
\noun{pnn}, is $\Omega\left(N^{2}\right)$ and indeed orders of magnitude
slower. These conclusions are corroborated by inspecting the NDC,
which depend on the accelerator used but not on the implementation
or the hardware, and are roughly correlated with the running time.

Overall, our results show that the \noun{pnns(init)} scheme is able
to improve the success rate of any \noun{init} algorithm at the cost
of a modest time penalty, and that it is a considerably better at
achieving this than the alternatives. It is particularly interesting
to consider the case of the \emph{Birch1} dataset, for which \noun{unif},
\noun{maxmin} and \noun{km++} all have 0\% success rate. Both \noun{gkm++}
and \noun{ref(km++)} are intended to improve \noun{km++} at the cost
of additional computations, and \noun{ref(gkm++)} combines the two
approaches; yet, none of them improves the success rate beyond 3\%.
The \noun{pnns(km++)} method, on the other hand, achieves 100\% success
rate on \emph{Birch1} in the same or smaller amount of time as those
other methods (and fewer NDCs).

\subsection{Real-world datasets\label{subsec:real-world}}

\begin{figure*}
\begin{centering}
\includegraphics[width=1\textwidth]{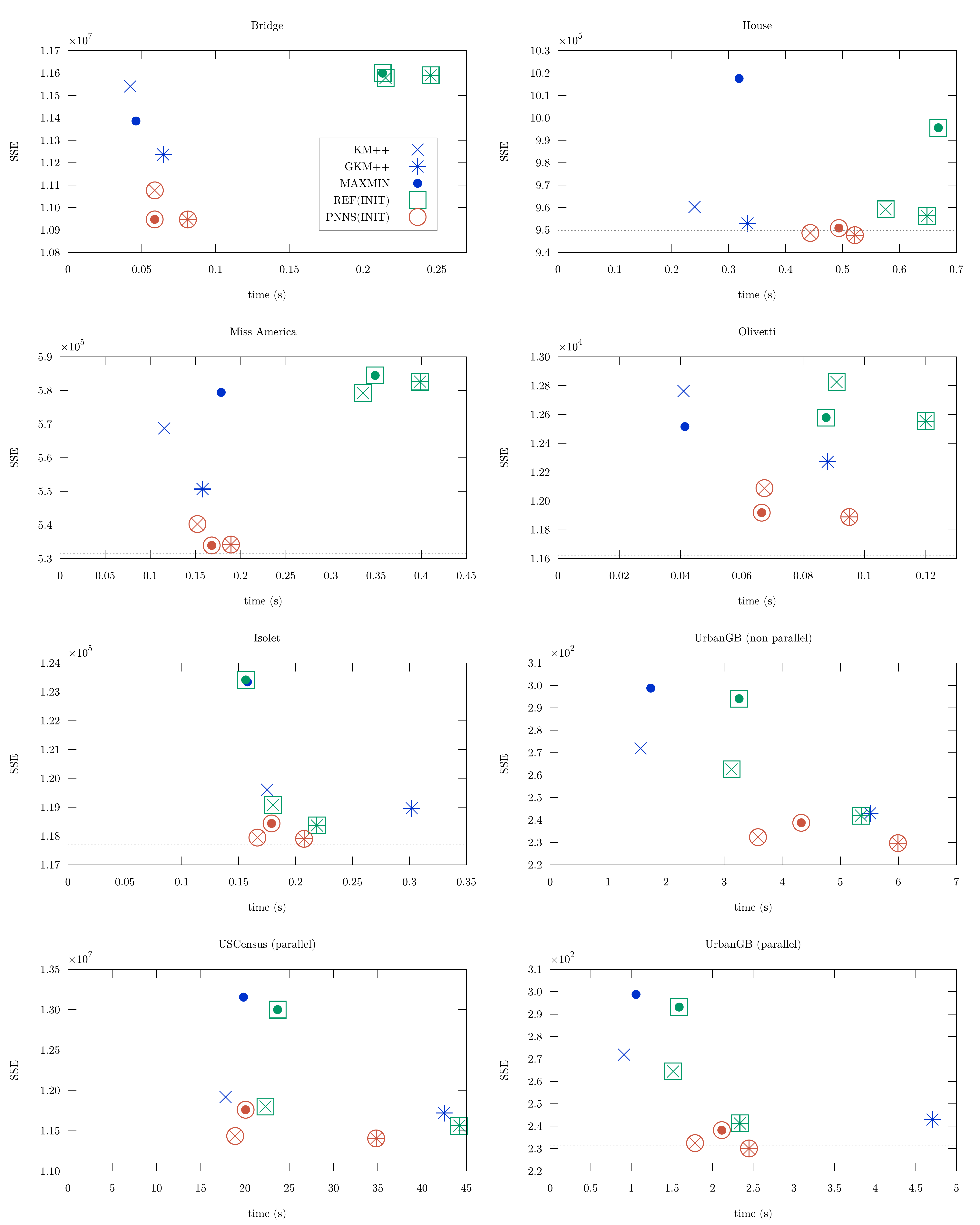}
\par\end{centering}
\caption{\label{fig:result2}$\mathrm{SSE}$ cost vs convergence time, averages
over $100$ samples (only $30$ for \emph{UrbanGB} and \emph{USCensus}),
for the real-world datasets. The meta-methods symbols enclose the
symbol for their \noun{init} algorithm. The dashed lines are the costs
achieved by \noun{pnn}, whose timings would be off-scale. The top
6 panels are non-parallel test; the bottom row are tests performed
with 4 threads in parallel.}
\end{figure*}

We tested 7 challenging real-world datasets (see table~\ref{tab:datasets}).
The first three, \emph{Bridge}, \emph{House} and \emph{Miss America},
were obtained
from the \href{http://cs.uef.fi/sipu/datasets/}{UEF repository}~\citep{franti2018k};
\emph{UrbanGB},\emph{ Isolet}
and \emph{USCensus} are from the
\href{http://archive.ics.uci.edu/ml}{UCI repository}~\citep{UCI-MachineLearningRepository}
and \emph{Olivetti} from~\href{https://scikit-learn.org}{scikit-learn}~\citep{scikit-learn}. All of these
are comparatively large and quite challenging for $\mathrm{SSE}$
optimization. The \emph{Isolet} dataset was chosen as the hardest
one among those tested in \citet{celebi2013comparative} (based
on the results reported there); the \emph{USCensus} dataset was chosen
for its large size, to provide a test of parallelization efficiency;
the other 5 datasets were also tested in \citet{baldassi2022recombinator},
where it was shown that even sophisticated, state-of-the-art evolutionary
algorithms cannot easily find their global minima (and indeed it is
not even clear if they can find them at all). We did not scale the
datasets, except \emph{UrbanGB} for which we scaled the longitude
by a factor of $1.7$ to make distances roughly proportional to geographical
distances, and \emph{USCensus} for which we linearly transformed each
dimension individually to make them fit into the range $\left[-1,1\right]$.
For all datasets, except \emph{UrbanGB} and \emph{Isolet}, the choice
of $k$ follows existing literature and is otherwise arbitrary.

We performed $100$ tests for each dataset ($30$ for \emph{UrbanGB}
and \emph{USCensus}) and computed the average and minimum $\mathrm{SSE}$
cost achieved across the runs (the latter metric can provide an indication
about what can be achieved with a multiple-restarts strategy by each
algorithm), as well as the average running time, NDCs, Lloyd's iterations.
As for the previous section, we report our most representative results
here; the full results can be found in Appendices~\ref{Asubsec:tables-real-nonp}
and~\ref{Asubsec:tables-real-para}. We first present the results
for the non-parallel case (on all datasets except \emph{USCensus}).
Some results are presented in table~\ref{tab:results2}, confirming
that the \noun{pnns} family of algorithms attains results superior
to all other linear algorithms, in comparable times (and comparable
NDC), both in terms of the average and of the minimum $\mathrm{SSE}$
cost. More specifically, \noun{pnns(init)} consistently achieves better
costs than both \noun{init} and \noun{ref(init)}, for all datasets
and all the tested \noun{init} (note that \noun{ref(init)} is not
consistently better than \noun{init}). This is confirmed by a statistical
significance analysis (standard two-sided Wilcoxon rank-sum test,
all $p$-values smaller than $10^{-11}$); in fact, in nearly all
cases the average cost for \noun{pnns(init)} is even smaller than
the minimum cost for \noun{init}. In many cases, the average time
for \noun{pnns(init)} is shorter than for \noun{ref(init)}. \noun{pnns(gkm++)}
generally achieves the best $\mathrm{SSE}$s (possibly on par with
other \noun{pnns} methods), and in 2 cases out of 6 it even outperforms
\noun{pnn}, which belongs to a different computational class and is
considerably slower.

In fig.~\ref{fig:result2}, we plot the mean $\mathrm{SSE}$ vs convergence
time for the best linear ``plain'' methods, \noun{km++}, \noun{gkm++}
and \noun{maxmin}, their \noun{ref(init)} versions\footnote{For the Olivetti dataset $N/k=10$ and thus we used $J=5$ instead
of the default $J=10$.}, and their \noun{pnns(init)} versions. They all have comparable timings,
but the \noun{pnns(init)} family is consistently below the others.
The dashed lines denote for reference the costs achieved by \noun{pnn}
(its timings would all be off-scale, cf. table~\ref{tab:results2}),
showing that even when it performs better than the algorithms in the
\noun{pnns} family, the difference is generally relatively small.

In order to test the effect of parallelization on the performance
of the algorithm, we also performed some tests with multi-threading
enabled, using $4$ threads. We tested the two largest datasets, \emph{USCensus}
and \emph{UrbanGB}. The results are shown in the last two rows of
fig.~\ref{fig:result2} (complete data in Appendix~\ref{Asubsec:tables-real-para});
the case of \emph{UrbanGB} allows a comparison with the non-parallel
ones; note that we did not run \noun{pnn} on \emph{USCensus} because
it would be impractical. The results are qualitatively similar to
the non-parallel versions.

As for the synthetic datasets, it is particularly interesting to compare
the different existing methods that can be used to improve on \noun{km++},
namely \noun{gkm++}, \noun{ref(km++)} and \noun{ref(gkm++)}, with
our method \noun{pnns(km++)}. Our method achieves better SSE costs
than all of the others; in 5 out of 7 datasets it is also faster (for
\emph{House} only \noun{gkm++} is faster, for \emph{UrbanGB} only
\noun{ref(km++)} is faster). Even in this real-world setting, our
\noun{pnns} scheme provides a better trade-off between cost and computational
time than the existing alternatives.

\section{Discussion\label{sec:discussion}}

We have presented a scheme for $k$-means seeding called PNN-smoothing
that can be applied to any existing linear ($O\left(kND\right)$)
algorithm, with a limited impact on the scaling and on the convergence
times. Our experiments, performed with an efficient implementation
and state-of-the-art techniques, show clear and consistent improvements
on challenging synthetic and real-world datasets, and systematically
superior results (both in terms of quality and of speed) with respect
to the similar ``refine'' smoothing scheme (also note that to the
best of our knowledge we were the first to report tests for \noun{ref(init)}
with \noun{init}$\ne$\noun{unif}, finding that it does not systematically
improve over \noun{init}). One particularly interesting case is that
of the very popular $k$-means++ seeding algorithm: we showed that
our scheme outperforms alternative enhancement techniques (i.e. making
it ``greedy'', using ``refine'', or both) in terms of both quality
and speed.\footnote{At the time of writing, greedy-$k$-means++ is the default seeding
method in the very popular scikit-learn package; according to our
results, \noun{pnns(km++)} would be a superior default.}

The overall picture is unchanged in a parallel multi-threading context.
We also verified that, as one would expect, the results are qualitatively
the same regardless of the scheme used to accelerate $k$-means iteration.
The experiments also indicate that PNN-smoothing is not particularly
susceptible to the original seeding scheme.

Overall, our results do not highlight any clear best among the \noun{pnns
}algorithms: \noun{pnns(km++)} is ususally the fastest, and it produces
consistently good results, but \noun{pnns(maxmin)} is not much different,
while the best costs are usually obtained by \noun{pnns(gkm++)}. When
trying to improve the cost sacrificing performance in difficult cases,
the best strategy is probably to use \noun{pnns(gkm++)} with a larger
$\rho$. From the results reported above, one can obtain an estimate
of the potential gain of increasing $\rho$ by looking at the difference
between the costs achieved with $\rho=1$ and those achieved with
\noun{pnn}. Since this difference is rather small,\footnote{Indeed, our results can also be summarized by stating that \noun{pnns}
can provide solutions of comparable quality to \noun{pnn} but with
a quasi-linear scaling rather than a quadratic one.} one would not expect a large improvement (in Appendix~\ref{Asec:tables}
we report the results with $\rho=10$). Another option would be to
apply the PNN-smoothing scheme recursively (e.g.~\noun{pnns(pnns(km++))}
and similar): we thoroughly explored this possibility,\footnote{We also considered a fully recursive version, which we call \noun{pnnsr},
that calls \noun{pnns} recursively until the subsets have size $N<2k$,
at which point it uses \noun{pnn}. It achieves good costs, but it
scales at least as $O\left(N\log N\right)$. See the associated code
for implementation details.} but concluded that it does not bring significant advantages over
simply using a larger $\rho$. On the other hand, if minimizing the
costs is of paramount importance, there exist better optimization
schemes than simple $k$-means, most notably evolutionary algorithms
that use $k$-means as their starting point \citep{franti2000genetic,baldassi2022recombinator};
for those, the seeding process is comparatively less important in
general, but it can still be crucial in some cases (as discussed in
ref.~\citep{baldassi2022recombinator}, when trying to access low-cost
regions of the configuration space, improving the seeding is usually
more efficient than blind exploration via random mutations). Assessing
the effect of better seeding on algorithmic schemes that go beyond
simple $k$-means (like the above-mentioned evolutionary schemes,
or for scenarios in which full Lloyd's iterations are too costly,
or data cannot fit into memory, etc.) is left for future work.

\bibliography{references}
\bibliographystyle{tmlr}

\appendix
\cleardoublepage{}

\section{\label{Asec:tables}Appendix: Complete numerical results}

In this section we report the full data for all numerical tests of
sec.~\ref{sec:experiments} of the main text. This complements tables
\ref{tab:results1} and \ref{tab:results2} of the main text, and
includes additional measures. When we report the average number of
Lloyd's iterations, we include the ones preformed during seeding by
\noun{refine(init)} and \noun{pnns(init)}, but we scale them by a
factor of $J$ to keep them comparable to the ones performed during
the final optimization.

Whenever we report means, we also report the standard deviations.

\subsection{\label{Asubsec:tables-synth}Synthetic datasets}

We tested $1000$ samples for \emph{A3}, \emph{Unbalance} and \emph{Dim1024},
and $100$ samples for \emph{Birch1} and \emph{Birch2}.

\subsubsection{Convergence time (in s)}

{\scriptsize{}}{\scriptsize\par}
\begin{center}
{\footnotesize{}}%
\begin{tabular}{llllll}
\hline 
{\footnotesize{}seeder} & \emph{\footnotesize{}A3}{\footnotesize{} }{\tiny{}($\times10^{-3}$)} & \emph{\footnotesize{}Birch1}{\footnotesize{} }{\tiny{}($\times10^{-1}$)} & \emph{\footnotesize{}Birch2}{\footnotesize{} }{\tiny{}($\times10^{-1}$)} & \emph{\footnotesize{}Unbalance}{\footnotesize{} }{\tiny{}($\times10^{-3}$)} & \emph{\footnotesize{}Dim1024}{\footnotesize{} }{\tiny{}($\times10^{-3}$)}\tabularnewline
\hline 
\noun{\footnotesize{}unif} & {\footnotesize{}$5.1\pm1.5$} & {\footnotesize{}$2.3\pm0.6$} & {\footnotesize{}$0.63\pm0.10$} & {\footnotesize{}$3.1\pm1.8$} & {\footnotesize{}$5.5\pm1.3$}\tabularnewline
\noun{\footnotesize{}maxmin} & {\footnotesize{}$4.0\pm0.9$} & {\footnotesize{}$2.0\pm0.4$} & {\footnotesize{}$0.75\pm0.10$} & {\footnotesize{}$0.7\pm0.8$} & {\footnotesize{}$5.4\pm0.8$}\tabularnewline
\noun{\footnotesize{}km++} & {\footnotesize{}$4.1\pm1.2$} & {\footnotesize{}$1.78\pm0.38$} & {\footnotesize{}$0.66\pm0.10$} & {\footnotesize{}$1.1\pm1.3$} & {\footnotesize{}$5.4\pm1.5$}\tabularnewline
\noun{\footnotesize{}gkm++} & {\footnotesize{}$6.5\pm1.4$} & {\footnotesize{}$2.44\pm0.26$} & {\footnotesize{}$1.64\pm0.11$} & {\footnotesize{}$0.9\pm1.0$} & {\footnotesize{}$12.8\pm1.7$}\tabularnewline
\noun{\footnotesize{}ref(unif)} & {\footnotesize{}$14.7\pm2.5$} & {\footnotesize{}$2.97\pm0.33$} & {\footnotesize{}$1.39\pm0.09$} & {\footnotesize{}$4.7\pm1.5$} & {\footnotesize{}$20.5\pm3.4$}\tabularnewline
\noun{\footnotesize{}ref(maxmin)} & {\footnotesize{}$10.2\pm2.5$} & {\footnotesize{}$2.40\pm0.21$} & {\footnotesize{}$1.25\pm0.11$} & {\footnotesize{}$1.5\pm1.1$} & {\footnotesize{}$13.8\pm2.8$}\tabularnewline
\noun{\footnotesize{}ref(km++)} & {\footnotesize{}$12.1\pm2.7$} & {\footnotesize{}$2.49\pm0.25$} & {\footnotesize{}$1.28\pm0.10$} & {\footnotesize{}$1.8\pm1.2$} & {\footnotesize{}$13.6\pm2.9$}\tabularnewline
\hline 
\end{tabular}{\footnotesize\par}
\par\end{center}

\subsubsection{SSE cost}
\begin{center}
{\footnotesize{}}%
\begin{tabular}{llllll}
\hline 
{\footnotesize{}seeder} & \emph{\footnotesize{}A3} & \emph{\footnotesize{}Birch1} & \emph{\footnotesize{}Birch2} & \emph{\footnotesize{}Unbalance} & \emph{\footnotesize{}Dim1024}\tabularnewline
\hline 
\noun{\footnotesize{}unif} & {\footnotesize{}$11.5\pm1.4$} & {\footnotesize{}$110\pm4$} & {\footnotesize{}$1.56\pm0.22$} & {\footnotesize{}$6.3\pm1.1$} & {\footnotesize{}$\left(1.4\pm0.4\right)\cdot10^{4}$}\tabularnewline
\noun{\footnotesize{}maxmin} & {\footnotesize{}$8.5\pm0.6$} & {\footnotesize{}$106.5\pm3.2$} & {\footnotesize{}$0.84\pm0.10$} & {\footnotesize{}$6\pm4$} & {\footnotesize{}$5.3938\pm0.0$}\tabularnewline
\noun{\footnotesize{}km++} & {\footnotesize{}$9.4\pm0.9$} & {\footnotesize{}$104.7\pm3.0$} & {\footnotesize{}$0.85\pm0.10$} & {\footnotesize{}$1.1\pm0.6$} & {\footnotesize{}$20\pm220$}\tabularnewline
\noun{\footnotesize{}gkm++} & {\footnotesize{}$7.7\pm0.5$} & {\footnotesize{}$99.5\pm2.7$} & {\footnotesize{}$0.528\pm0.035$} & {\footnotesize{}$0.68\pm0.16$} & {\footnotesize{}$5.3938\pm0.0$}\tabularnewline
\noun{\footnotesize{}ref(unif)} & {\footnotesize{}$9.7\pm0.8$} & {\footnotesize{}$104.0\pm2.8$} & {\footnotesize{}$1.22\pm0.11$} & {\footnotesize{}$4.5\pm1.3$} & {\footnotesize{}$\left(5.3\pm2.4\right)\cdot10^{3}$}\tabularnewline
\noun{\footnotesize{}ref(maxmin)} & {\footnotesize{}$7.17\pm0.28$} & {\footnotesize{}$100.4\pm1.9$} & {\footnotesize{}$0.56\pm0.05$} & {\footnotesize{}$0.646935\pm0.0$} & {\footnotesize{}$5.3938\pm0.0$}\tabularnewline
\noun{\footnotesize{}ref(km++)} & {\footnotesize{}$8.4\pm0.5$} & {\footnotesize{}$101.0\pm2.5$} & {\footnotesize{}$0.71\pm0.05$} & {\footnotesize{}$0.648\pm0.030$} & {\footnotesize{}$5.3938\pm0.0$}\tabularnewline
\hline 
\end{tabular}{\footnotesize\par}
\par\end{center}

\subsubsection{CI (average)}
\begin{center}
{\scriptsize{}}%
\begin{tabular}{llllll}
\hline 
{\scriptsize{}seeder} & \emph{\scriptsize{}A3} & \emph{\scriptsize{}Birch1} & \emph{\scriptsize{}Birch2} & \emph{\scriptsize{}Unbalance} & \emph{\scriptsize{}Dim1024}\tabularnewline
\hline 
\noun{\scriptsize{}unif} & {\scriptsize{}$6.6\pm1.6$} & {\scriptsize{}$6.7\pm1.7$} & {\scriptsize{}$16.4\pm2.5$} & {\scriptsize{}$3.93\pm0.35$} & {\scriptsize{}$3.7\pm1.1$}\tabularnewline
\noun{\scriptsize{}maxmin} & {\scriptsize{}$2.9\pm1.0$} & {\scriptsize{}$5.6\pm1.3$} & {\scriptsize{}$7.4\pm1.8$} & {\scriptsize{}$0.9\pm0.6$} & {\scriptsize{}$0$}\tabularnewline
\noun{\scriptsize{}km++} & {\scriptsize{}$4.1\pm1.2$} & {\scriptsize{}$4.8\pm1.2$} & {\scriptsize{}$7.1\pm1.6$} & {\scriptsize{}$0.5\pm0.6$} & {\scriptsize{}$0.004\pm0.063$}\tabularnewline
\noun{\scriptsize{}gkm++} & {\scriptsize{}$1.7\pm0.8$} & {\scriptsize{}$2.7\pm1.1$} & {\scriptsize{}$1.6\pm0.8$} & {\scriptsize{}$0.05\pm0.23$} & {\scriptsize{}$0$}\tabularnewline
\noun{\scriptsize{}ref(unif)} & {\scriptsize{}$4.7\pm1.1$} & {\scriptsize{}$4.5\pm1.1$} & {\scriptsize{}$12.4\pm1.8$} & {\scriptsize{}$3.4\pm0.8$} & {\scriptsize{}$1.4\pm0.7$}\tabularnewline
\noun{\scriptsize{}ref(maxmin)} & {\scriptsize{}$0.8\pm0.5$} & {\scriptsize{}$3.1\pm0.8$} & {\scriptsize{}$2.4\pm1.0$} & {\scriptsize{}$0$} & {\scriptsize{}$0$}\tabularnewline
\noun{\scriptsize{}ref(km++)} & {\scriptsize{}$2.8\pm0.8$} & {\scriptsize{}$3.3\pm1.0$} & {\scriptsize{}$4.9\pm1.0$} & {\scriptsize{}$0.00\pm0.04$} & {\scriptsize{}$0$}\tabularnewline
\hline 
\end{tabular}{\scriptsize\par}
\par\end{center}

\subsubsection{Success rate}

{\scriptsize{}}{\scriptsize\par}
\begin{center}
{\footnotesize{}}%
\begin{tabular}{lrrrrr}
\hline 
{\footnotesize{}seeder} & \emph{\footnotesize{}A3} & \emph{\footnotesize{}Birch1} & \emph{\footnotesize{}Birch2} & \emph{\footnotesize{}Unbalance} & \emph{\footnotesize{}Dim1024}\tabularnewline
\hline 
\noun{\footnotesize{}unif} & {\footnotesize{}$0$} & {\footnotesize{}$0$} & {\footnotesize{}$0$} & {\footnotesize{}$0$} & {\footnotesize{}$0.001$}\tabularnewline
\noun{\footnotesize{}maxmin} & {\footnotesize{}$0.004$} & {\footnotesize{}$0$} & {\footnotesize{}$0$} & {\footnotesize{}$0.223$} & {\footnotesize{}$1$}\tabularnewline
\noun{\footnotesize{}km++} & {\footnotesize{}$0$} & {\footnotesize{}$0$} & {\footnotesize{}$0$} & {\footnotesize{}$0.541$} & {\footnotesize{}$0.996$}\tabularnewline
\noun{\footnotesize{}gkm++} & {\footnotesize{}$0.051$} & {\footnotesize{}$0.02$} & {\footnotesize{}$0.06$} & {\footnotesize{}$0.946$} & {\footnotesize{}$1$}\tabularnewline
\noun{\footnotesize{}ref(unif)} & {\footnotesize{}$0$} & {\footnotesize{}$0$} & {\footnotesize{}$0$} & {\footnotesize{}$0$} & {\footnotesize{}$0.055$}\tabularnewline
\noun{\footnotesize{}ref(maxmin)} & {\footnotesize{}$0.254$} & {\footnotesize{}$0$} & {\footnotesize{}$0.01$} & {\footnotesize{}$1$} & {\footnotesize{}$1$}\tabularnewline
\noun{\footnotesize{}ref(km++)} & {\footnotesize{}$0.002$} & {\footnotesize{}$0$} & {\footnotesize{}$0$} & {\footnotesize{}$0.998$} & {\footnotesize{}$1$}\tabularnewline
\hline 
\end{tabular}{\footnotesize\par}
\par\end{center}

\subsubsection{Normalized distance computations}
\begin{center}
{\footnotesize{}}%
\begin{tabular}{llllll}
\hline 
{\footnotesize{}seeder} & \emph{\footnotesize{}A3} & \emph{\footnotesize{}Birch1} & \emph{\footnotesize{}Birch2} & \emph{\footnotesize{}Unbalance} & \emph{\footnotesize{}Dim1024}\tabularnewline
\hline 
\noun{\footnotesize{}unif} & {\footnotesize{}$1.80\pm0.18$} & {\footnotesize{}$2.52\pm0.37$} & {\footnotesize{}$1.146\pm0.027$} & {\footnotesize{}$2.7\pm0.6$} & {\footnotesize{}$1.89\pm0.32$}\tabularnewline
\noun{\footnotesize{}maxmin} & {\footnotesize{}$1.33\pm0.09$} & {\footnotesize{}$2.08\pm0.29$} & {\footnotesize{}$1.073\pm0.020$} & {\footnotesize{}$1.26\pm0.10$} & {\footnotesize{}$1.07812\pm0.0$}\tabularnewline
\noun{\footnotesize{}km++} & {\footnotesize{}$1.41\pm0.12$} & {\footnotesize{}$2.00\pm0.25$} & {\footnotesize{}$1.063\pm0.014$} & {\footnotesize{}$1.35\pm0.31$} & {\footnotesize{}$1.079\pm0.009$}\tabularnewline
\noun{\footnotesize{}gkm++} & {\footnotesize{}$5.08\pm0.05$} & {\footnotesize{}$6.46\pm0.16$} & {\footnotesize{}$5.972\pm0.004$} & {\footnotesize{}$3.78\pm0.11$} & {\footnotesize{}$3.89062\pm0.0$}\tabularnewline
\noun{\footnotesize{}ref(unif)} & {\footnotesize{}$3.94\pm0.12$} & {\footnotesize{}$3.75\pm0.21$} & {\footnotesize{}$2.393\pm0.019$} & {\footnotesize{}$4.6\pm0.5$} & {\footnotesize{}$6.6\pm0.4$}\tabularnewline
\noun{\footnotesize{}ref(maxmin)} & {\footnotesize{}$3.139\pm0.038$} & {\footnotesize{}$3.08\pm0.11$} & {\footnotesize{}$2.207\pm0.008$} & {\footnotesize{}$2.468\pm0.019$} & {\footnotesize{}$4.1136\pm0.0015$}\tabularnewline
\noun{\footnotesize{}ref(km++)} & {\footnotesize{}$3.39\pm0.07$} & {\footnotesize{}$3.23\pm0.14$} & {\footnotesize{}$2.240\pm0.012$} & {\footnotesize{}$2.61\pm0.08$} & {\footnotesize{}$4.1140\pm0.0023$}\tabularnewline
\hline 
\end{tabular}{\footnotesize\par}
\par\end{center}

\subsubsection{Lloyd's iterations}
\begin{center}
{\footnotesize{}}%
\begin{tabular}{llllll}
\hline 
{\footnotesize{}seeder} & \emph{\footnotesize{}A3} & \emph{\footnotesize{}Birch1} & \emph{\footnotesize{}Birch2} & \emph{\footnotesize{}Unbalance} & \emph{\footnotesize{}Dim1024}\tabularnewline
\hline 
\noun{\footnotesize{}unif} & {\footnotesize{}$26\pm10$} & {\footnotesize{}$120\pm40$} & {\footnotesize{}$45\pm12$} & {\footnotesize{}$33\pm15$} & {\footnotesize{}$2.7\pm1.3$}\tabularnewline
\noun{\footnotesize{}maxmin} & {\footnotesize{}$19\pm10$} & {\footnotesize{}$95\pm36$} & {\footnotesize{}$41\pm11$} & {\footnotesize{}$3.2\pm1.1$} & {\footnotesize{}$1\pm0$}\tabularnewline
\noun{\footnotesize{}km++} & {\footnotesize{}$21\pm9$} & {\footnotesize{}$85\pm31$} & {\footnotesize{}$41\pm13$} & {\footnotesize{}$11\pm13$} & {\footnotesize{}$1.00\pm0.07$}\tabularnewline
\noun{\footnotesize{}gkm++} & {\footnotesize{}$13\pm7$} & {\footnotesize{}$60\pm23$} & {\footnotesize{}$23\pm11$} & {\footnotesize{}$3\pm6$} & {\footnotesize{}$1\pm0$}\tabularnewline
\noun{\footnotesize{}ref(unif)} & {\footnotesize{}$26\pm13$} & {\footnotesize{}$100\pm40$} & {\footnotesize{}$56\pm18$} & {\footnotesize{}$32\pm20$} & {\footnotesize{}$4.6\pm2.9$}\tabularnewline
\noun{\footnotesize{}ref(maxmin)} & {\footnotesize{}$14\pm10$} & {\footnotesize{}$68\pm34$} & {\footnotesize{}$35\pm19$} & {\footnotesize{}$2.3\pm0.6$} & {\footnotesize{}$2.001\pm0.031$}\tabularnewline
\noun{\footnotesize{}ref(km++)} & {\footnotesize{}$21\pm11$} & {\footnotesize{}$73\pm37$} & {\footnotesize{}$46\pm19$} & {\footnotesize{}$5\pm4$} & {\footnotesize{}$2.001\pm0.031$}\tabularnewline
\hline 
\end{tabular}{\scriptsize{}\newpage}{\scriptsize\par}
\par\end{center}

\subsection{\label{Asubsec:tables-real-nonp}Real-world datasets (non-parallel)}

We tested $100$ samples for each dataset ($30$ for \emph{UrbanGB}).

\subsubsection{Convergence time (in s)}
\begin{center}
{\footnotesize{}}%
\begin{tabular}{lllllll}
\hline 
{\footnotesize{}seeder} & \emph{\footnotesize{}Bridge}{\footnotesize{} }{\tiny{}($\times10^{-2}$)} & \emph{\footnotesize{}House}{\footnotesize{} }{\tiny{}($\times10^{-1}$)} & \emph{\footnotesize{}Miss A.}{\footnotesize{} }{\tiny{}($\times10^{-1}$)} & \emph{\footnotesize{}Urb.GB} & \emph{\footnotesize{}Olivetti}{\footnotesize{} }{\tiny{}($\times10^{-2}$)} & \emph{\footnotesize{}Isolet}{\footnotesize{} }{\tiny{}($\times10^{-1}$)}\tabularnewline
\hline 
\noun{\footnotesize{}unif} & {\footnotesize{}$4.17\pm0.38$} & {\footnotesize{}$3.3\pm0.6$} & {\footnotesize{}$1.13\pm0.09$} & {\footnotesize{}$1.63\pm0.23$} & {\footnotesize{}$2.7\pm0.6$} & {\footnotesize{}$1.52\pm0.30$}\tabularnewline
\noun{\footnotesize{}maxmin} & {\footnotesize{}$4.6\pm0.5$} & {\footnotesize{}$3.2\pm0.5$} & {\footnotesize{}$1.78\pm0.33$} & {\footnotesize{}$1.73\pm0.20$} & {\footnotesize{}$4.1\pm0.6$} & {\footnotesize{}$1.58\pm0.21$}\tabularnewline
\noun{\footnotesize{}km++} & {\footnotesize{}$4.22\pm0.36$} & {\footnotesize{}$2.40\pm0.37$} & {\footnotesize{}$1.15\pm0.11$} & {\footnotesize{}$1.56\pm0.13$} & {\footnotesize{}$4.10\pm0.39$} & {\footnotesize{}$1.75\pm0.29$}\tabularnewline
\noun{\footnotesize{}gkm++} & {\footnotesize{}$6.45\pm0.35$} & {\footnotesize{}$3.33\pm0.33$} & {\footnotesize{}$1.58\pm0.13$} & {\footnotesize{}$5.51\pm0.16$} & {\footnotesize{}$8.8\pm0.6$} & {\footnotesize{}$3.02\pm0.29$}\tabularnewline
\noun{\footnotesize{}ref(unif)} & {\footnotesize{}$22.0\pm0.5$} & {\footnotesize{}$6.5\pm0.4$} & {\footnotesize{}$3.26\pm0.11$} & {\footnotesize{}$3.65\pm0.28$} & {\footnotesize{}$8.0\pm0.5$} & {\footnotesize{}$1.68\pm0.24$}\tabularnewline
\noun{\footnotesize{}ref(maxmin)} & {\footnotesize{}$21.3\pm0.6$} & {\footnotesize{}$6.7\pm0.5$} & {\footnotesize{}$3.49\pm0.17$} & {\footnotesize{}$3.26\pm0.24$} & {\footnotesize{}$8.8\pm0.5$} & {\footnotesize{}$1.56\pm0.22$}\tabularnewline
\hline 
\end{tabular}{\footnotesize\par}
\par\end{center}

\subsubsection{SSE cost (average)}
\begin{center}
{\footnotesize{}}%
\begin{tabular}{lllllll}
\hline 
{\footnotesize{}seeder} & \emph{\footnotesize{}Bridge}{\footnotesize{} }{\tiny{}($\times10^{7}$)} & \emph{\footnotesize{}House}{\footnotesize{} }{\tiny{}($\times10^{5}$)} & \emph{\footnotesize{}Miss A.}{\footnotesize{} }{\tiny{}($\times10^{5}$)} & \emph{\footnotesize{}Urb.GB}{\footnotesize{} }{\tiny{}($\times10^{2}$)} & \emph{\footnotesize{}Olivetti}{\footnotesize{} }{\tiny{}($\times10^{4}$)} & \emph{\footnotesize{}Isolet}{\footnotesize{} }{\tiny{}($\times10^{5}$)}\tabularnewline
\hline 
\noun{\footnotesize{}unif} & {\footnotesize{}$1.178\pm0.009$} & {\footnotesize{}$10.11\pm0.13$} & {\footnotesize{}$6.07\pm0.05$} & {\footnotesize{}$6.8\pm1.1$} & {\footnotesize{}$1.296\pm0.027$} & {\footnotesize{}$1.196\pm0.010$}\tabularnewline
\noun{\footnotesize{}maxmin} & {\footnotesize{}$1.139\pm0.005$} & {\footnotesize{}$10.18\pm0.07$} & {\footnotesize{}$5.794\pm0.028$} & {\footnotesize{}$2.99\pm0.07$} & {\footnotesize{}$1.252\pm0.015$} & {\footnotesize{}$1.233\pm0.014$}\tabularnewline
\noun{\footnotesize{}km++} & {\footnotesize{}$1.154\pm0.007$} & {\footnotesize{}$9.60\pm0.04$} & {\footnotesize{}$5.687\pm0.037$} & {\footnotesize{}$2.72\pm0.06$} & {\footnotesize{}$1.276\pm0.021$} & {\footnotesize{}$1.196\pm0.010$}\tabularnewline
\noun{\footnotesize{}gkm++} & {\footnotesize{}$1.124\pm0.004$} & {\footnotesize{}$9.529\pm0.027$} & {\footnotesize{}$5.507\pm0.016$} & {\footnotesize{}$2.430\pm0.020$} & {\footnotesize{}$1.227\pm0.013$} & {\footnotesize{}$1.190\pm0.007$}\tabularnewline
\noun{\footnotesize{}ref(unif)} & {\footnotesize{}$1.162\pm0.006$} & {\footnotesize{}$9.95\pm0.10$} & {\footnotesize{}$5.850\pm0.035$} & {\footnotesize{}$5.3\pm0.4$} & {\footnotesize{}$1.298\pm0.023$} & {\footnotesize{}$1.189\pm0.006$}\tabularnewline
\hline 
\end{tabular}{\footnotesize\par}
\par\end{center}

\subsubsection{SSE cost (minimum)}
\begin{center}
{\footnotesize{}}%
\begin{tabular}{lrrrrrr}
\hline 
{\footnotesize{}seeder} & \emph{\footnotesize{}Bridge}{\footnotesize{} }{\tiny{}($\times10^{7}$)} & \emph{\footnotesize{}House}{\footnotesize{} }{\tiny{}($\times10^{5}$)} & \emph{\footnotesize{}Miss A.}{\footnotesize{} }{\tiny{}($\times10^{5}$)} & \emph{\footnotesize{}Urb.GB}{\footnotesize{} }{\tiny{}($\times10^{2}$)} & \emph{\footnotesize{}Olivetti}{\footnotesize{} }{\tiny{}($\times10^{4}$)} & \emph{\footnotesize{}Isolet}{\footnotesize{} }{\tiny{}($\times10^{5}$)}\tabularnewline
\hline 
\noun{\footnotesize{}unif} & {\footnotesize{}$1.15723$} & {\footnotesize{}$9.87713$} & {\footnotesize{}$5.95536$} & {\footnotesize{}$5.42031$} & {\footnotesize{}$1.23747$} & {\footnotesize{}$1.1795$}\tabularnewline
\noun{\footnotesize{}maxmin} & {\footnotesize{}$1.12869$} & {\footnotesize{}$10.0159$} & {\footnotesize{}$5.72134$} & {\footnotesize{}$2.90322$} & {\footnotesize{}$1.21735$} & {\footnotesize{}$1.20639$}\tabularnewline
\noun{\footnotesize{}km++} & {\footnotesize{}$1.14219$} & {\footnotesize{}$9.52638$} & {\footnotesize{}$5.59949$} & {\footnotesize{}$2.60788$} & {\footnotesize{}$1.23035$} & {\footnotesize{}$1.18025$}\tabularnewline
\noun{\footnotesize{}gkm++} & {\footnotesize{}$1.11421$} & {\footnotesize{}$9.45992$} & {\footnotesize{}$5.47583$} & {\footnotesize{}$2.37869$} & {\footnotesize{}$1.19908$} & {\footnotesize{}$1.17709$}\tabularnewline
\noun{\footnotesize{}ref(unif)} & {\footnotesize{}$1.14562$} & {\footnotesize{}$9.77128$} & {\footnotesize{}$5.76604$} & {\footnotesize{}$4.39285$} & {\footnotesize{}$1.24007$} & {\footnotesize{}$1.17688$}\tabularnewline
\noun{\footnotesize{}ref(maxmin)} & {\footnotesize{}$1.14719$} & {\footnotesize{}$9.81462$} & {\footnotesize{}$5.7451$} & {\footnotesize{}$2.84408$} & {\footnotesize{}$1.22378$} & {\footnotesize{}$1.20994$}\tabularnewline
\hline 
\end{tabular}{\footnotesize\par}
\par\end{center}

\subsubsection{Normalized distance computations}
\begin{center}
{\footnotesize{}}%
\begin{tabular}{lllllll}
\hline 
{\footnotesize{}seeder} & \emph{\footnotesize{}Bridge} & \emph{\footnotesize{}House} & \emph{\footnotesize{}Miss A.} & \emph{\footnotesize{}Urb.GB} & \emph{\footnotesize{}Olivetti} & \emph{\footnotesize{}Isolet}\tabularnewline
\hline 
\noun{\footnotesize{}unif} & {\footnotesize{}$8.4\pm0.5$} & {\footnotesize{}$5.7\pm0.9$} & {\footnotesize{}$13.5\pm0.9$} & {\footnotesize{}$1.60\pm0.23$} & {\footnotesize{}$1.79\pm0.14$} & {\footnotesize{}$3.35\pm0.36$}\tabularnewline
\noun{\footnotesize{}maxmin} & {\footnotesize{}$8.9\pm0.8$} & {\footnotesize{}$4.5\pm0.6$} & {\footnotesize{}$23\pm4$} & {\footnotesize{}$1.196\pm0.037$} & {\footnotesize{}$2.21\pm0.07$} & {\footnotesize{}$3.14\pm0.25$}\tabularnewline
\noun{\footnotesize{}km++} & {\footnotesize{}$8.2\pm0.6$} & {\footnotesize{}$3.71\pm0.39$} & {\footnotesize{}$13.7\pm1.3$} & {\footnotesize{}$1.229\pm0.029$} & {\footnotesize{}$2.26\pm0.07$} & {\footnotesize{}$3.36\pm0.35$}\tabularnewline
\noun{\footnotesize{}gkm++} & {\footnotesize{}$13.2\pm0.5$} & {\footnotesize{}$9.27\pm0.32$} & {\footnotesize{}$19.5\pm1.5$} & {\footnotesize{}$8.168\pm0.033$} & {\footnotesize{}$6.04\pm0.05$} & {\footnotesize{}$6.92\pm0.33$}\tabularnewline
\noun{\footnotesize{}ref(unif)} & {\footnotesize{}$44.8\pm0.8$} & {\footnotesize{}$8.3\pm0.6$} & {\footnotesize{}$39.4\pm1.0$} & {\footnotesize{}$3.15\pm0.15$} & {\footnotesize{}$5.68\pm0.10$} & {\footnotesize{}$4.62\pm0.26$}\tabularnewline
\noun{\footnotesize{}ref(maxmin)} & {\footnotesize{}$43.5\pm0.8$} & {\footnotesize{}$7.4\pm0.5$} & {\footnotesize{}$42.2\pm1.9$} & {\footnotesize{}$2.509\pm0.029$} & {\footnotesize{}$6.40\pm0.08$} & {\footnotesize{}$4.48\pm0.21$}\tabularnewline
\noun{\footnotesize{}ref(km++)} & {\footnotesize{}$43.8\pm0.7$} & {\footnotesize{}$6.78\pm0.39$} & {\footnotesize{}$40.5\pm1.3$} & {\footnotesize{}$2.62\pm0.08$} & {\footnotesize{}$6.53\pm0.10$} & {\footnotesize{}$4.75\pm0.25$}\tabularnewline
\hline 
\end{tabular}{\footnotesize\par}
\par\end{center}

\subsubsection{Lloyd's iterations}
\begin{center}
{\footnotesize{}}%
\begin{tabular}{lllllll}
\hline 
{\footnotesize{}seeder} & \emph{\footnotesize{}Bridge} & \emph{\footnotesize{}House} & \emph{\footnotesize{}Miss A.} & \emph{\footnotesize{}Urb.GB} & \emph{\footnotesize{}Olivetti} & \emph{\footnotesize{}Isolet}\tabularnewline
\hline 
\noun{\footnotesize{}unif} & {\footnotesize{}$21\pm5$} & {\footnotesize{}$124\pm29$} & {\footnotesize{}$31\pm7$} & {\footnotesize{}$120\pm28$} & {\footnotesize{}$8.1\pm2.6$} & {\footnotesize{}$38\pm14$}\tabularnewline
\noun{\footnotesize{}maxmin} & {\footnotesize{}$25\pm6$} & {\footnotesize{}$134\pm35$} & {\footnotesize{}$62\pm17$} & {\footnotesize{}$96\pm29$} & {\footnotesize{}$8.1\pm2.6$} & {\footnotesize{}$28\pm9$}\tabularnewline
\noun{\footnotesize{}km++} & {\footnotesize{}$22\pm5$} & {\footnotesize{}$100\pm26$} & {\footnotesize{}$36\pm7$} & {\footnotesize{}$84\pm17$} & {\footnotesize{}$8.1\pm2.4$} & {\footnotesize{}$38\pm14$}\tabularnewline
\noun{\footnotesize{}gkm++} & {\footnotesize{}$20\pm4$} & {\footnotesize{}$90\pm25$} & {\footnotesize{}$37\pm9$} & {\footnotesize{}$86\pm25$} & {\footnotesize{}$6.1\pm1.5$} & {\footnotesize{}$34\pm12$}\tabularnewline
\noun{\footnotesize{}ref(unif)} & {\footnotesize{}$26\pm10$} & {\footnotesize{}$116\pm27$} & {\footnotesize{}$40\pm13$} & {\footnotesize{}$130\pm60$} & {\footnotesize{}$10\pm4$} & {\footnotesize{}$37\pm17$}\tabularnewline
\noun{\footnotesize{}ref(maxmin)} & {\footnotesize{}$27\pm11$} & {\footnotesize{}$140\pm40$} & {\footnotesize{}$53\pm18$} & {\footnotesize{}$100\pm60$} & {\footnotesize{}$9\pm4$} & {\footnotesize{}$30\pm15$}\tabularnewline
\noun{\footnotesize{}ref(km++)} & {\footnotesize{}$25\pm10$} & {\footnotesize{}$104\pm30$} & {\footnotesize{}$44\pm15$} & {\footnotesize{}$110\pm50$} & {\footnotesize{}$9\pm4$} & {\footnotesize{}$36\pm17$}\tabularnewline
\hline 
\end{tabular}{\scriptsize{}\newpage}{\scriptsize\par}
\par\end{center}

\subsection{\label{Asubsec:tables-real-para}Real-world datasets (parallel, 4
threads)}

We tested $30$ samples for each dataset.

\subsubsection{Convergence time (in s)}
\begin{center}
{\footnotesize{}}%
\begin{tabular}{lll}
\hline 
{\footnotesize{}seeder} & \emph{\footnotesize{}Urb.GB} & \emph{\footnotesize{}USCensus}\tabularnewline
\hline 
\noun{\footnotesize{}unif} & {\footnotesize{}$0.86\pm0.13$} & {\footnotesize{}$13.2\pm3.3$}\tabularnewline
\noun{\footnotesize{}maxmin} & {\footnotesize{}$1.06\pm0.12$} & {\footnotesize{}$19.8\pm2.1$}\tabularnewline
\noun{\footnotesize{}km++} & {\footnotesize{}$0.91\pm0.13$} & {\footnotesize{}$17.8\pm2.2$}\tabularnewline
\noun{\footnotesize{}gkm++} & {\footnotesize{}$4.71\pm0.17$} & {\footnotesize{}$42.5\pm3.0$}\tabularnewline
\noun{\footnotesize{}ref(unif)} & {\footnotesize{}$1.74\pm0.12$} & {\footnotesize{}$17.1\pm2.0$}\tabularnewline
\noun{\footnotesize{}ref(maxmin)} & {\footnotesize{}$1.59\pm0.19$} & {\footnotesize{}$23.7\pm2.1$}\tabularnewline
\noun{\footnotesize{}ref(km++)} & {\footnotesize{}$1.51\pm0.16$} & {\footnotesize{}$22.3\pm2.2$}\tabularnewline
\hline 
\end{tabular}{\footnotesize\par}
\par\end{center}

\subsubsection{SSE cost (average)}
\begin{center}
{\footnotesize{}}%
\begin{tabular}{lll}
\hline 
{\footnotesize{}seeder} & \emph{\footnotesize{}Urb.GB}{\footnotesize{} }{\tiny{}($\times10^{2}$)} & \emph{\footnotesize{}USCensus}{\footnotesize{} }{\tiny{}($\times10^{7}$)}\tabularnewline
\hline 
\noun{\footnotesize{}unif} & {\footnotesize{}$6.8\pm1.1$} & {\footnotesize{}$1.200\pm0.010$}\tabularnewline
\noun{\footnotesize{}maxmin} & {\footnotesize{}$2.99\pm0.07$} & {\footnotesize{}$1.32\pm0.04$}\tabularnewline
\noun{\footnotesize{}km++} & {\footnotesize{}$2.72\pm0.05$} & {\footnotesize{}$1.192\pm0.011$}\tabularnewline
\noun{\footnotesize{}gkm++} & {\footnotesize{}$2.430\pm0.020$} & {\footnotesize{}$1.172\pm0.007$}\tabularnewline
\noun{\footnotesize{}ref(unif)} & {\footnotesize{}$5.26\pm0.35$} & {\footnotesize{}$1.181\pm0.008$}\tabularnewline
\noun{\footnotesize{}ref(maxmin)} & {\footnotesize{}$2.93\pm0.05$} & {\footnotesize{}$1.300\pm0.032$}\tabularnewline
\noun{\footnotesize{}ref(km++)} & {\footnotesize{}$2.64\pm0.05$} & {\footnotesize{}$1.180\pm0.006$}\tabularnewline
\hline 
\end{tabular}{\footnotesize\par}
\par\end{center}

\subsubsection{SSE cost (minimum)}
\begin{center}
{\footnotesize{}}%
\begin{tabular}{lrr}
\hline 
{\footnotesize{}seeder} & \emph{\footnotesize{}Urb.GB}{\footnotesize{} }{\tiny{}($\times10^{2}$)} & \emph{\footnotesize{}USCensus}{\footnotesize{} }{\tiny{}($\times10^{7}$)}\tabularnewline
\hline 
\noun{\footnotesize{}unif} & {\footnotesize{}$5.42031$} & {\footnotesize{}$1.18059$}\tabularnewline
\noun{\footnotesize{}maxmin} & {\footnotesize{}$2.90322$} & {\footnotesize{}$1.25624$}\tabularnewline
\noun{\footnotesize{}km++} & {\footnotesize{}$2.61092$} & {\footnotesize{}$1.17627$}\tabularnewline
\noun{\footnotesize{}gkm++} & {\footnotesize{}$2.37869$} & {\footnotesize{}$1.15802$}\tabularnewline
\noun{\footnotesize{}ref(unif)} & {\footnotesize{}$4.65076$} & {\footnotesize{}$1.16404$}\tabularnewline
\noun{\footnotesize{}ref(maxmin)} & {\footnotesize{}$2.83593$} & {\footnotesize{}$1.2423$}\tabularnewline
\noun{\footnotesize{}ref(km++)} & {\footnotesize{}$2.50803$} & {\footnotesize{}$1.16902$}\tabularnewline
\hline 
\end{tabular}{\footnotesize\par}
\par\end{center}
\end{document}